\pdfoutput=1
\documentclass[10pt]{article}

\usepackage[letterpaper,textwidth=5.5in,textheight=9in,centering]{geometry}
\usepackage{times}
\usepackage{natbib}
\setcitestyle{authoryear,round,citesep={;},aysep={,},yysep={;}}

\makeatletter
\renewcommand{\normalsize}{\@setfontsize\normalsize\@xpt{11}}
\renewcommand{\small}{\@setfontsize\small\@ixpt{10}}
\renewcommand{\footnotesize}{\@setfontsize\footnotesize\@ixpt{10}}
\renewcommand{\scriptsize}{\@setfontsize\scriptsize\@viipt{8}}
\renewcommand{\tiny}{\@setfontsize\tiny\@vipt{7}}
\renewcommand{\large}{\@setfontsize\large\@xiipt{14}}
\renewcommand{\Large}{\@setfontsize\Large\@xivpt{16}}
\renewcommand{\LARGE}{\@setfontsize\LARGE\@xviipt{20}}
\renewcommand{\huge}{\@setfontsize\huge\@xxpt{23}}
\renewcommand{\Huge}{\@setfontsize\Huge\@xxvpt{28}}
\normalsize
\def\@listi{\leftmargin\leftmargini}
\def\@listii{\leftmargin\leftmarginii
  \labelwidth\leftmarginii\advance\labelwidth-\labelsep
  \topsep 2pt plus 1pt minus 0.5pt
  \parsep 1pt plus 0.5pt minus 0.5pt
  \itemsep \parsep}
\def\@listiii{\leftmargin\leftmarginiii
  \labelwidth\leftmarginiii\advance\labelwidth-\labelsep
  \topsep 1pt plus 0.5pt minus 0.5pt
  \parsep \z@ \partopsep 0.5pt plus 0pt minus 0.5pt
  \itemsep \topsep}
\makeatother

\renewcommand{\footnoterule}{\kern-3pt \hrule width 12pc \kern 2.6pt}

\makeatletter
\renewcommand{\section}{\@startsection{section}{1}{\z@}%
  {-2.0ex \@plus -0.5ex \@minus -.2ex}{1.5ex \@plus 0.3ex \@minus 0.2ex}%
  {\normalfont\Large\bfseries}}
\renewcommand{\subsection}{\@startsection{subsection}{2}{\z@}%
  {-1.8ex \@plus -0.5ex \@minus -.2ex}{0.8ex \@plus .2ex}%
  {\normalfont\large\bfseries}}
\renewcommand{\subsubsection}{\@startsection{subsubsection}{3}{\z@}%
  {-1.5ex \@plus -0.5ex \@minus -.2ex}{0.5ex \@plus .2ex}%
  {\normalfont\normalsize\bfseries}}
\renewcommand{\paragraph}{\@startsection{paragraph}{4}{\z@}%
  {1.5ex \@plus .5ex \@minus .2ex}{-1em}%
  {\normalfont\normalsize\bfseries}}
\makeatother

\makeatletter
\renewcommand{\@maketitle}{%
  \newpage
  \begin{center}
    \def\@makefnmark{\@textsuperscript{\normalfont\@thefnmark}}%
    {\LARGE \@title \par}%
    \vskip 1em%
    {\large \lineskip .5em \begin{tabular}[t]{c}\@author\end{tabular}\par}%
  \end{center}%
  \par\vskip 0.5em}
\makeatother
\usepackage{amsmath,amsfonts,bm}

\def\eqref#1{equation~\ref{#1}}

\def\1{\bm{1}}

\def\vg{{\bm{g}}}

\def\vx{{\bm{x}}}
\def\vy{{\bm{y}}}

\def\mA{{\bm{A}}}

\def\mC{{\bm{C}}}
\def\mD{{\bm{D}}}
\def\mE{{\bm{E}}}

\def\mG{{\bm{G}}}
\def\mH{{\bm{H}}}
\def\mI{{\bm{I}}}
\def\mJ{{\bm{J}}}

\def\mL{{\bm{L}}}
\def\mM{{\bm{M}}}

\def\mO{{\bm{O}}}
\def\mP{{\bm{P}}}
\def\mQ{{\bm{Q}}}
\def\mR{{\bm{R}}}
\def\mS{{\bm{S}}}

\def\mU{{\bm{U}}}
\def\mV{{\bm{V}}}
\def\mW{{\bm{W}}}
\def\mX{{\bm{X}}}
\def\mY{{\bm{Y}}}
\def\mZ{{\bm{Z}}}

\def\mLambda{{\bm{\Lambda}}}
\def\mSigma{{\bm{\Sigma}}}

\DeclareMathAlphabet{\mathsfit}{\encodingdefault}{\sfdefault}{m}{sl}
\SetMathAlphabet{\mathsfit}{bold}{\encodingdefault}{\sfdefault}{bx}{n}

\usepackage{url}

\usepackage{amsmath}
\usepackage{amssymb}
\usepackage{amsthm}
\usepackage{bm}
\usepackage{graphicx}
\usepackage{subcaption}
\usepackage{xspace}
\usepackage{xcolor}
\usepackage{booktabs}
\usepackage{array}
\usepackage{threeparttable}
\usepackage{algorithm}
\usepackage{algpseudocode}
\usepackage{hyperref}
\usepackage{etoc}
\hypersetup{hidelinks,
  pdftitle={Dobb: Shard-Local Preconditioning in Block-Diagonal Eigenbases},
  pdfauthor={Shi-Yuk Fong, Shuhua Yu, Jiawei Zhang}}
\newcommand{\dobb}{Dobb\xspace}
\newcommand{\dobbG}{Dobb-G\xspace}
\newcommand{\dobbLtwo}{Dobb-\texorpdfstring{$\ell_2$}{l2}\xspace}
\newcommand{\dobbGLtwo}{Dobb-G-\texorpdfstring{$\ell_2$}{l2}\xspace}
\newcommand{\DP}{\ensuremath{\mathrm{DP}}}
\newcommand{\FS}{\ensuremath{\mathrm{FS}}}
\newcommand{\TP}{\ensuremath{\mathrm{TP}}}

\title{Dobb: Shard-Local Preconditioning in Block-Diagonal Eigenbases}

\author{%
Shi-Yuk Fong\textsuperscript{1}\quad
Shuhua Yu\textsuperscript{2}\thanks{Co-last and corresponding authors.}\quad
Jiawei Zhang\textsuperscript{1}\footnotemark[1]\\[0.5ex]
\normalsize\textsuperscript{1}Department of Computer Sciences, University of Wisconsin--Madison, Madison, WI, USA\\
\normalsize\textsuperscript{2}Meta FAIR, Menlo Park, CA, USA\\[0.5ex]
\normalsize\textsuperscript{1}\texttt{\fontencoding{T1}\selectfont\{syfong, jiaweizhang\}@cs.wisc.edu}, \textsuperscript{2}\texttt{yu.shuhuaxh@gmail.com}
}

\theoremstyle{plain}

\theoremstyle{definition}

\begin{document}

\maketitle
\begin{abstract}
Training language models whose parameters and optimizer state exceed a single node's memory requires sharding these tensors across devices.
Under tensor parallelism (TP) or fully sharded data parallelism (FSDP), matrix-aware optimizers such as Muon introduce communication to perform global matrix operations across shards.
We propose \dobb (\textbf{D}istributed \textbf{O}ptimizer with \textbf{B}lock-diagonal Eigen\textbf{b}ases), a matrix-aware optimizer designed to compute updates locally within parameter shards.
\dobb approximates the input-activation second moment with diagonal blocks and tracks a complete basis within each block.
It projects momentum into these bases, applies sign normalization to the coefficients, and reconstructs the update.
When each block lies within one shard under TP or FSDP, \dobb applies its matrix update locally, requiring no communication beyond data-parallel synchronization of the activation second moments. 
We benchmark \dobb against Dion, a distributed matrix-aware optimizer designed to retain the benefits of orthonormalized updates while reducing communication.
On GPT-2 XL (1.56B parameters), \dobb with 16 blocks reaches a validation loss of 2.476, compared with 2.511 for Dion at rank fraction 1/16.
Under FSDP on 32 H100 GPUs, \dobb reduces the amortized optimizer-call time from 124.1~ms for Dion to 9.2~ms.
\end{abstract}

\begin{figure}[H]
\centering
\captionsetup{font=small,skip=4pt}
\includegraphics[width=0.55\linewidth]{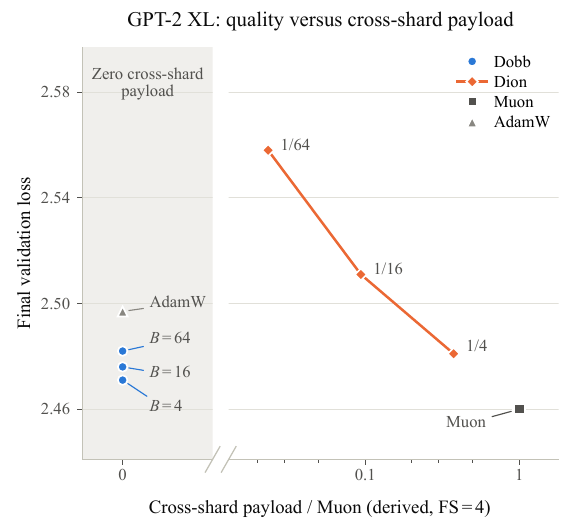}
\caption{GPT-2 XL (1.56B parameters) validation loss and optimizer-specific cross-shard payload.
Losses are measured under the shared update-scale protocol (Table~\ref{tab:quality}).
Labels give block count $B$ or rank fraction $\rho$.
For a weight matrix of size $m\times n$, optimizer-specific cross-shard payload is the logical number of scalar entries
exchanged between devices holding different shards of that matrix during one optimizer step; collective factors and bytes
per entry are omitted.
We express this derived count relative to Muon's dense count $mn$.
At Dion rank $r$, the per-matrix ratio is $r(m+n)/(mn)$; the aggregate ratio for the plotted GPT-2 XL matrices is
$1.5\rho$.
Payloads are derived for FSDP with $\FS=4$, where each \dobb block lies within one shard; \dobb and AdamW then have zero payload under this definition.
The count excludes forward-backward communication and \dobb's separately counted data-parallel synchronization of activation second moments (Section~\ref{sec:comm}).
}
\label{fig:teaser}
\end{figure}

\etocdepthtag.toc{main}
\section{Introduction}
\label{sec:intro}

Language-model scaling increases the memory requirements for parameters, gradients, and optimizer state \citep{kaplan2020scaling,rajbhandari2020zero}.
When these quantities exceed a single GPU's capacity, training distributes them across devices through tensor parallelism (TP) or fully sharded data parallelism (FSDP) \citep{shoeybi2019megatron,zhao2023fsdp}.
Under either layout, each device holds one part of every sharded weight matrix, and we call this part a shard; data-parallel replicas hold copies of the same shards.
Cross-shard communication is communication among the devices that hold different shards of one matrix.
These layouts make matrix-aware optimization a distributed problem: an update defined on a complete weight matrix may require information held by several devices.
Muon orthogonalizes momentum \citep{jordan2024muon}, which introduces dense optimizer-specific communication in conventional sharded implementations \citep{ahn2025dion,khaled2025muonbp}.

MuonBP reduces communication by performing local Newton--Schulz orthogonalization between periodic global steps, but those global steps still require full cross-shard communication \citep{khaled2025muonbp}.
The frequency of global orthogonalization controls the tradeoff between communication cost and optimization quality.
Another approach, Dion, reduces the size of each exchange by approximating momentum with low-rank factors and retaining the residual through error feedback \citep{ahn2025dion}.
Its optimizer step communicates these factors across shards, so communication scales with the update rank.

In this work, we introduce \dobb (\textbf{D}istributed \textbf{O}ptimizer with \textbf{B}lock-diagonal Eigen\textbf{b}ases) to reduce optimizer communication.
We approximate the activation second moment by a block-diagonal matrix and retain a complete eigenbasis within every block.
This construction discards cross-block correlations without imposing a reduced-rank constraint on the update.
When these blocks align with shard boundaries, the optimizer step requires no cross-shard communication beyond the data-parallel synchronization of the activation second moments.

Dion derives its low-rank factors from momentum; we instead project momentum onto an input-side basis from an exponentially averaged activation second moment, normalize the coefficients, and reconstruct the update.
For a mini-batch gradient, projection into its own activation eigenbasis separates input spectral scales from the output-gradient structure, leaving the latter in the coefficients (Section~\ref{sec:background}).
In our GPT-2 Small experiments, activation second moments exhibit greater spectral concentration than output-gradient second moments in most matched matrices (Figure~\ref{fig:rank}).
This motivates using the activation eigenbasis to expose the more concentrated factor's directional scaling for normalization (Appendix~\ref{app:basis}).

Both sign and column normalization remove positive input spectral scales from a mini-batch gradient expressed in its own activation eigenbasis.
We choose the element-wise sign, which additionally removes magnitude differences within each coefficient column.
In our synthetic comparison, this choice produces more orthogonal updates than column normalization except when the output-gradient spectrum is nearly flat (Table~\ref{tab:synthetic-orthogonality}).
Sign normalization also avoids the norm reduction required by column normalization when the output axis is sharded, because it acts on each coefficient independently.

We make four contributions:
\begin{itemize}
    \item We introduce \dobb, which normalizes momentum coefficients in block-diagonal activation second-moment eigenbases while retaining every direction within each block.
    \item \dobb requires only that each block lie within one shard.
    Under this condition, the optimizer step introduces zero optimizer-specific cross-shard bytes beyond the data-parallel synchronization of the activation second moments.
    \item We compare \dobb with Muon and Dion on GPT-2-style models up to 1.56B parameters under a shared update-scale protocol.
    Across these scales, \dobb exhibits less validation-loss degradation under block partitioning than Dion under corresponding rank reductions.
    \item We benchmark \dobb against Muon and Dion under FSDP and tensor parallelism on GPT-2-style models up to 8.1B parameters and on Qwen3 models up to 32B.
    In every measured configuration, \dobb's optimizer call is shorter than those of Muon and Dion (Section~\ref{sec:results-fsdp}).
\end{itemize}

\section{Related work}
\label{sec:related}

\paragraph{Matrix-aware optimization.}
Matrix-aware optimizers use the intrinsic two-dimensional structure of weight matrices to construct parameter updates rather than treating each entry independently.
Muon orthogonalizes momentum using Newton--Schulz iterations, while RMNP uses row normalization \citep{jordan2024muon,deng2026rmnp}.
K-FAC builds a Kronecker-factored approximation to layerwise curvature \citep{martens2015kfac}.
Shampoo accumulates gradient-derived matrix second moments, and SOAP applies Adam in their eigenbases \citep{gupta2018shampoo,vyas2024soap}.

\paragraph{Sharding and tensor parallelism.}
ZeRO reduces memory redundancy across data-parallel workers by progressively sharding optimizer state, gradients, and parameters across its three stages \citep{rajbhandari2020zero}.
FSDP shards all three and gathers parameters as needed for forward and backward computation \citep{zhao2023fsdp}.
FSDP2 uses per-parameter sharding, which facilitates composition with tensor parallelism \citep{liang2025torchtitan}.
Tensor parallelism, as used in Megatron-LM, partitions weight matrices and their matrix multiplications across devices \citep{shoeybi2019megatron}.
The resulting parameter layout determines which operands are local and which require communication during the optimizer step.

\paragraph{Distributed matrix updates.}
While element-wise optimizer updates parallelize naturally, matrix-aware updates introduce dependencies among optimizer state, matrix computations, and the underlying parameter layout.
Dion retains a full-size momentum buffer distributed across shards and reduces communication through low-rank factor operations and error feedback, with communication scaling with the update rank \citep{ahn2025dion}.
Distributed Shampoo takes a different approach.
It distributes preconditioner state and computation by parameter block within data-parallel process groups, then all-gathers the computed search directions within each group at every optimizer step \citep{shi2023distshampoo}.
MatrixFSDP takes a systems approach under FSDP: it places each whole matrix on one owner rank, so Muon runs unchanged without optimizer-step collectives, and it leaves matrices sharded by tensor parallelism outside its scope \citep{gao2026matrixfsdp}.

\section{Gradient structure and motivation}
\label{sec:background}

We seek a basis for normalizing momentum coefficients while keeping the optimizer step local to parameter shards.
We motivate the basis source, normalization, and block partition through gradient structure, spectral comparisons, and the distributed parameter layout.

\paragraph{A mini-batch gradient has two sources of structure.}
Consider a layer $\vy=\mW\vx$ with $\mW\in\mathbb{R}^{m\times n}$.
For the mini-batch $\mathcal{B}_t$ at optimizer step $t$, containing $N$ contributing examples or token positions, let $f_{\mathcal{B}_t}$ denote its mean loss and let $\vg_i\in\mathbb{R}^{m}$ denote the output gradient associated with its $i$th contribution before the $1/N$ averaging.
With inputs $\vx_i\in\mathbb{R}^{n}$, the layer's stochastic gradient is
\begin{equation}
\label{eq:batch-gradient}
\nabla_{\mW}f_{\mathcal{B}_t}(\mW_t)
 = \frac{1}{N}\sum_{i=1}^{N}\vg_i\vx_i^\top
 = \frac{1}{N}\mD_t\mX_t^\top,
\end{equation}
where $\mX_t=[\vx_1,\ldots,\vx_N]\in\mathbb{R}^{n\times N}$ and $\mD_t=[\vg_1,\ldots,\vg_N]\in\mathbb{R}^{m\times N}$.
  Equation~\ref{eq:batch-gradient} shows that every row of the mini-batch gradient is a combination of the inputs $\vx_i$,
  while every column is a combination of the output gradients $\vg_i$.
  K-FAC approximates a layerwise Fisher block by a Kronecker product of the second moments of these two sources
  \citep{martens2015kfac}.
  For the current mini-batch, the corresponding uncentered moments are
  \begin{equation}
  \label{eq:factors}
  \mA_t=\frac{1}{N}\mX_t\mX_t^\top \in\mathbb{R}^{n\times n},
  \qquad
  \mG_t=\frac{1}{N}\mD_t\mD_t^\top \in\mathbb{R}^{m\times m}.
  \end{equation}
  The nonzero eigenspaces of $\mA_t$ and $\mG_t$ span the observed input and output-gradient directions, respectively,
  providing candidate bases for the two axes of a matrix update.
  Neither factor requires materializing per-example weight gradients.
\paragraph{A common template: project, normalize, reconstruct.}
Several matrix-aware updates project the momentum $\mM\in\mathbb{R}^{m\times n}$ onto an orthonormal basis $\mU$, normalize the coefficients with a function $\mathcal{N}$, and reconstruct:
\begin{equation}
\label{eq:template}
\mO=\mathcal{N}(\mM\mU)\,\mU^\top,
\end{equation}
or apply the same form to $\mM^\top$.
Muon approximates the polar factor of the momentum with five Newton--Schulz iterations \citep{jordan2024muon}.
With the compact singular value decomposition $\mM=\mV\mSigma\mU^\top$, the coefficients are $\mM\mU=\mV\mSigma$, and normalizing each coefficient column to unit $\ell_2$ norm gives $\mV$.
The polar factor $\mV\mU^\top$ is therefore Equation~\ref{eq:template} with the right singular vectors of $\mM$ as the basis and column normalization as $\mathcal{N}$.
Dion applies Equation~\ref{eq:template} to $\mM^\top$ using a rank-$r$ orthonormal basis $\mV_r\in\mathbb R^{m\times r}$, with $\mathcal{N}$ being column normalization as well. $\mV_r$ is obtained by one warm-started power-iteration step on $\mM$, approximating its leading left singular subspace.
Dion retains the rank-$r$ approximation residual in its momentum through error feedback \citep{ahn2025dion}.

For weight gradients $\mH_t=\nabla_{\mW}f_{\mathcal{B}_t}(\mW)$, Shampoo adds $\mH_t\mH_t^\top$ and $\mH_t^\top\mH_t$ to its left and right Gram matrices $\mL$ and $\mR$, respectively, and uses inverse powers of the accumulated matrices to precondition the update \citep{gupta2018shampoo}.
Let $\mQ_L$ and $\mQ_R$ contain the eigenvectors of the accumulated $\mL$ and $\mR$, so that $\mQ_L^\top\mH_t\mQ_R$ expresses the gradient in their two-sided coordinates.
SOAP applies Adam to gradients in these coordinates, while SPlus applies the element-wise sign to momentum in these coordinates \citep{vyas2024soap,frans2025splus}.
These weight-gradient Gram matrices differ from the activation and output-gradient second moments $\mA_t$ and $\mG_t$ defined above.
\citet{gong2026aro} analyze a general form of Equation~\ref{eq:template} as rotated steepest descent, $\mJ\psi(\mJ^\top\mM)$ with a rotation $\mJ$ and a base function $\psi$.
Their base functions include the sign and row normalization (the counterpart of column normalization for a basis applied on the left), and their unified view lists one-sided simplifications of SOAP and SPlus.
\dobb takes its basis from the eigenvectors of the activation second moment and, instead of column normalization, applies the element-wise sign, $\mathcal{N}(\mZ)=\operatorname{sign}(\mZ)/\sqrt{m}$.
We first compare the two basis sources, then assess sign against column normalization in the selected basis.

\begin{figure}[t]
\centering
\captionsetup[subfigure]{font=small,labelfont=normalfont,textfont=normalfont,skip=2pt}
\begin{subfigure}[t]{0.45\linewidth}
\centering
\includegraphics[width=\linewidth]{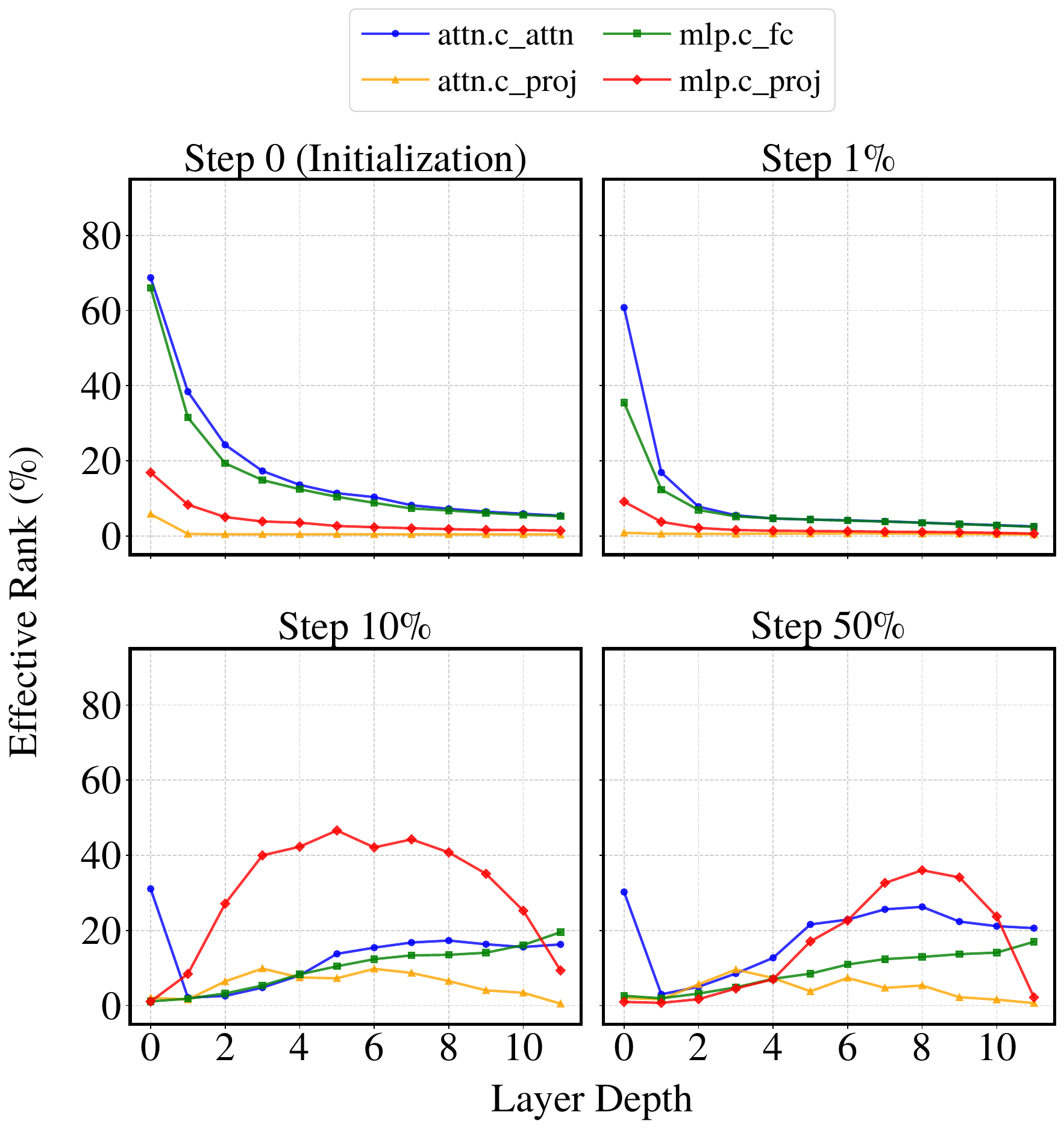}
\caption{Input-activation second moment $\mA_t$}
\end{subfigure}\hfill
\begin{subfigure}[t]{0.45\linewidth}
\centering
\includegraphics[width=\linewidth]{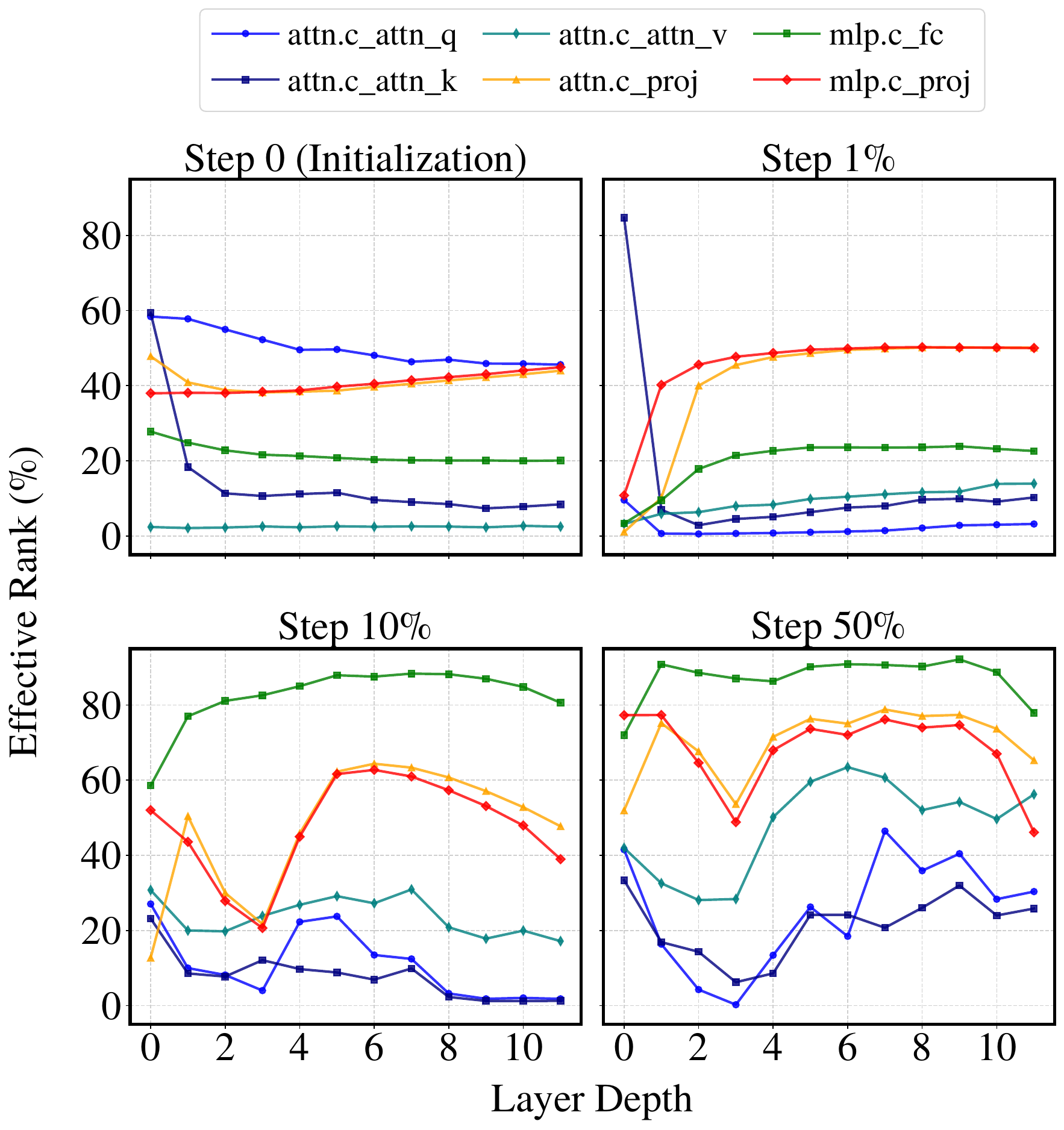}
\caption{Output-gradient second moment $\mG_t$}
\end{subfigure}
\caption{Effective rank of the second moments from GPT-2 Small runs on OpenWebText.
Within each plot, four panels show transformer layer depth at the indicated fraction of training completed; curves distinguish weight-matrix types.
From 50\% to 100\% of training, the effective-rank distribution remains stable.
Effective rank is $\exp(-\sum_i p_i\log p_i)$ as a percentage of the factor dimension, $n$ for $\mA_t$ and $m$ for $\mG_t$, where $p_i$ is the $i$th normalized eigenvalue.
The second-moment matrices are averaged over accumulated mini-batches.
The output-gradient plot separates the query, key, and value blocks of the fused attention projection.
}
\label{fig:rank}
\end{figure}

\paragraph{Spectral concentration motivates the input-side basis.}
In transformer training, weight gradients \citep{shen2025muonconv} and updates from SGD with momentum and Adam \citep{jordan2024muon} typically have low-rank structure. Muon approximately orthogonalizes its updates, making their singular values nearly equal, and achieves model quality comparable to AdamW with less compute \citep{jordan2024muon,liu2025muonscalable}.
We compare activation and output-gradient second moments using entropy-based effective rank, normalized by the factor dimension, across layer types and training checkpoints (Figure~\ref{fig:rank}).
A lower effective rank indicates that the spectrum is concentrated in fewer eigendirections. 
Figure~\ref{fig:rank} shows that activation factors exhibit greater spectral concentration than the corresponding output-gradient factors.
At 50\% of training, over the 72 weight matrices of GPT-2 Small with the query, key, and value projections counted separately, the median effective rank is 13\% for the activation second moment and 55\% for the output-gradient second moment, and the output-gradient factor has the higher effective rank in 65 of the 72 pairs.
\citet{shen2025muonconv} report highly concentrated singular values in the input feature matrices of MNIST and CIFAR-10 and in RoBERTa text embeddings.
An eigenbasis exposes spectral concentration as unequal scales along its coordinate directions.
This motivates using the activation eigenbasis to remove the more concentrated factor's directional scaling through sign normalization.
To see how the activation eigenbasis interacts with sign normalization, we express the mini-batch gradient in this basis.
Write the compact singular value decomposition as $\mX_t^\top=\mV_X\mSigma_X\mU^\top$, where the columns of $\mU$ are the eigenvectors of $\mA_t$ with positive eigenvalues.
The gradient coefficients in this basis are
\[
\nabla_{\mW}f_{\mathcal{B}_t}(\mW_t)\,\mU
=\frac{1}{N}\mD_t\mV_X\mSigma_X.
\]
Each diagonal entry of $\mSigma_X$ positively scales one column of these coefficients.
Since positive scaling does not change an entry's sign,
\[
\operatorname{sign}\!\left(
\nabla_{\mW}f_{\mathcal{B}_t}(\mW_t)\,\mU
\right)
=\operatorname{sign}(\mD_t\mV_X).
\]
Thus, sign normalization removes the activation singular-value scaling from the gradient coefficients.
The cancellation of activation singular-value scaling is exact for a mini-batch gradient expressed in the eigenbasis of that mini-batch's activation second moment.
For momentum expressed in a reused eigenbasis of an averaged activation second moment, this cancellation serves as motivation rather than an exact identity.
In Table~\ref{tab:synthetic-orthogonality}(a), the sign of a synthetic matrix with higher effective rank is more orthogonal. The basis ablation provides empirical support for this choice: the activation-side basis gives the lower final validation loss (Appendix~\ref{app:basis}).
We therefore use the eigenvectors of the activation second moment as coordinates for normalizing momentum coefficients.

\paragraph{Sign or column normalization.}
A column-normalized variant keeps \dobb's basis and applies Muon's column normalization to the same coefficients, so comparing it with \dobb isolates the normalization.
Column normalization also ignores positive column scaling and therefore acts on $\mD_t \mV_X$.
In the synthetic gradients of Table~\ref{tab:synthetic-orthogonality}(b), the columns of $\mD_t \mV_X$ have nearly equal norms, so column normalization changes the coefficients little and keeps the concentration of the output-gradient factor.
In these gradients, the sign gives the more orthogonal update unless the output-gradient second moment is nearly flat, with effective rank above about 75\% of its dimension.
In the GPT-2 Small runs of Figure~\ref{fig:rank}, 52 of the 72 output-gradient second moments have effective rank below 75\% at 50\% of training.
In training, sign normalization gives the lower final loss in all sixteen matched comparisons from GPT-2 Small to XL, by 0.008 to 0.026 (Appendix~\ref{app:norm}).
The sign also acts on each coefficient independently, whereas column normalization requires a reduction when the output axis is sharded.

\begin{table}[t]
\centering
\small
\caption{Orthogonality of updates on synthetic $768\times768$ gradients.
(a) A matrix $\mY$ with prescribed effective rank and its sign.
(b) Sign normalization (\dobb) and column normalization in the same eigenbasis of the full activation second moment (a single block), on gradients $\mD\mX^\top$ whose output-gradient second moment $\mG$ has the stated effective rank, with the effective rank of the activation second moment fixed at 15\%.
Effective rank is the entropy-based measure of Figure~\ref{fig:rank} as a percentage of the dimension, computed from the singular values of $\mY$ in (a).
Orthogonality is $\|\mO\|_*/(\sqrt{\min(m,n)}\,\|\mO\|_F)$ for a nonzero $m\times n$ matrix $\mO$, where $\|\cdot\|_*$ and $\|\cdot\|_F$ denote the nuclear and Frobenius norms; it equals 1 exactly when all singular values are equal.
Appendix~\ref{app:synthetic} describes the construction.
}
\label{tab:synthetic-orthogonality}
\begin{tabular}[t]{rrr}
\multicolumn{3}{c}{(a) Sign of a matrix} \\
\toprule
 & \multicolumn{2}{c}{Orthogonality} \\
\cmidrule(lr){2-3}
Effective rank of $\mY$ & $\mY$ & $\operatorname{sign}(\mY)$ \\
\midrule
80\% & 0.736 & 0.775 \\
50\% & 0.412 & 0.643 \\
20\% & 0.205 & 0.543 \\
5\% & 0.109 & 0.457 \\
\bottomrule
\end{tabular}
\hspace{1.5em}
\begin{tabular}[t]{rrr}
\multicolumn{3}{c}{(b) Normalization in the activation eigenbasis} \\
\toprule
 & \multicolumn{2}{c}{Orthogonality} \\
\cmidrule(lr){2-3}
Effective rank of $\mG$ & Sign (\dobb) & Column $\ell_2$ \\
\midrule
88\% & 0.854 & 0.864 \\
80\% & 0.846 & 0.849 \\
72\% & 0.839 & 0.834 \\
55\% & 0.820 & 0.796 \\
38\% & 0.796 & 0.748 \\
22\% & 0.762 & 0.680 \\
\bottomrule
\end{tabular}
\end{table}

\paragraph{Temporal stability motivates basis reuse.}
K-FAC exploits slowly changing second-moment factors and relatively stable curvature to amortize inverse computations across optimizer steps \citep[Appendix~G]{martens2015kfac}.
Following this rationale, we reuse the activation-side basis between periodic refreshes, reducing the amortized cost of basis computation.
Appendix~\ref{app:refresh} evaluates the resulting tradeoff between iteration time and optimization quality.

\paragraph{Sharding motivates a block-diagonal basis.}
Under TP and FSDP alike, each rank holds one shard of a weight matrix and the matching shards of its gradient and optimizer state, so a matrix-aware update of the complete matrix needs operands from other shards.
We therefore seek a basis that each rank can construct and apply from its own operands: a block-diagonal basis whose blocks align with shard boundaries.
We write $\TP$ for the tensor-parallel degree, $\FS$ for the FSDP shard degree, and $B$ for the number of second-moment blocks.
In the Megatron-LM layout, a column-parallel layer shards the output axis $m$ and receives a replicated input, whereas a row-parallel layer shards the input axis $n$ and reduces partial outputs \citep{shoeybi2019megatron}.
For row-parallel layers, we partition the activation second moment into diagonal blocks within the input shards, so each block basis acts on local momentum columns.
For column-parallel layers, the replicated input allows each rank to construct the same block bases and apply them to its local momentum rows.
Under FSDP, our implementation shards each matrix that \dobb updates along its input axis, so each rank holds whole blocks and their momentum columns, as in a row-parallel layer.
This construction accommodates both sharding axes under TP and the input-axis sharding under FSDP without requiring cross-shard communication during the optimizer step beyond the data-parallel synchronization of the activation second moments.
This synchronization spans FSDP ranks and data-parallel replicas, which process different data, but not the ranks of a TP group, which share their data (Section~\ref{sec:comm}).

\paragraph{Block partitioning retains a complete basis.}
The activation spectrum motivates the choice of basis, while sharding under TP and FSDP constrains how we construct and apply it.
Dion reduces communication through low-rank updates, restricting each update to a lower-dimensional subspace while retaining the approximation residual through error feedback \citep{ahn2025dion}.
A rank-$r$ update has at least $\min(m,n)-r$ zero singular values, which limits its orthogonality over the full matrix space even when its factors have orthonormal columns.
We instead partition the activation second moment into $B$ blocks and retain all eigenvectors within each block.
This construction discards cross-block correlations without imposing a reduced-rank constraint on the update and allows basis operations to remain local when blocks align with shard boundaries.
At $B=n$, each block holds one input coordinate, each block basis is $\pm1$, and the update reduces to $\operatorname{sign}(\mM)/\sqrt{m}$, sign descent with momentum \citep{bernstein2018signsgd}.
As $B$ increases from 1 to $n$, \dobb's update therefore moves from normalization in a single activation eigenbasis toward sign descent with momentum, whereas reducing Dion's rank increases the number of zero singular values in its update.

\section{Distributed Optimizer with Block-diagonal Eigenbases}
\label{sec:method}

\dobb normalizes momentum in a second-moment eigenbasis whose blocks can be assigned to individual shards.
All eigenvectors within each block are retained; $B$ controls the block partition.
Algorithm~\ref{alg:dobb} states \dobb for one weight matrix.

\paragraph{An exponentially averaged block-diagonal second moment.}
Partition the input coordinates into $B$ equal-sized contiguous sets $\mathcal{S}_1,\ldots,\mathcal{S}_B$.
For simplicity, we consider only block counts $B$ that divide $n$, so every block has width $n/B$; we leave partitions with unequal block widths, needed when $B$ does not divide $n$, to future work.
On each capture window, let $\mX_t^{\mathrm{cap}}\in\mathbb{R}^{n\times N_c}$ stack the inputs at the $N_c$ captured token positions and form
\begin{equation}
\label{eq:block-moment}
\begin{gathered}
\mS_{t,b}=\frac{1}{N_c}\mX^{\mathrm{cap}}_{t,\mathcal{S}_b,:}(\mX^{\mathrm{cap}}_{t,\mathcal{S}_b,:})^{\top},
\qquad
\mC_{t,b}=\beta_C\mC_{\mathrm{prev},b}+\mS_{t,b}.
\end{gathered}
\end{equation}
The subscript $\mathcal{S}_b,:$ selects the rows indexed by $\mathcal{S}_b$, and $:,\mathcal{S}_b$ selects the columns.
The accumulator $\mC_{\mathrm{prev},b}$ is the value from the previous capture, and $\beta_C$ is its decay.
The first capture initializes $\mC_{t,b}=\mS_{t,b}$.
This is an exponentially weighted second moment; multiplying the entire accumulator by a positive scalar leaves its eigenvectors unchanged.
We refresh the orthonormal basis $\mU_{t,b}$ every $T$ optimizer steps and reuse it between refreshes.
Although each block basis spans the full block space, its orientation changes the coefficients to which the element-wise sign is applied and thus the reconstructed update.
Each refresh applies one step of orthogonal iteration, initialized from the previous basis, to approximate the evolving eigenbasis.
Appendix~\ref{app:algorithm} states the refresh with Householder QR and with Cholesky QR, and Appendix~\ref{app:cholqr} compares the two routines in GPT-2 training.
The capture schedule and basis refresh are coupled, with $T=9$ as the principal setting.

\paragraph{Projection, normalization, and reconstruction.}
Each block projects its columns of the momentum $\mM_t$ of Algorithm~\ref{alg:dobb} onto its basis, normalizes the coefficients, and reconstructs the corresponding columns of the update $\mO_t$:
\begin{equation}
\label{eq:dobb-update}
\mZ_{t,b}=\mM_{t,:,\,\mathcal{S}_b}\mU_{t,b},
\qquad
\mO_{t,:,\,\mathcal{S}_b}
 =\mathcal{N}(\mZ_{t,b})\mU_{t,b}^{\top}.
\end{equation}
With the element-wise sign normalization $\mathcal{N}$ of Section~\ref{sec:background}, Equation~\ref{eq:dobb-update} is Equation~\ref{eq:template} with the block-diagonal basis whose diagonal blocks are $\mU_{t,1},\ldots,\mU_{t,B}$.

\begin{algorithm}[t]
\caption{\dobb for one weight matrix of a single layer.
Appendix~\ref{app:algorithm} gives the capture policy and the synchronization of statistics, and Algorithm~\ref{alg:dobb-full} gives the complete algorithm.}
\label{alg:dobb}
\begin{algorithmic}[1]
\Require $\mW\in\mathbb{R}^{m\times n}$; blocks $\{\mathcal{S}_b\}_{b=1}^{B}$; period $T$; decays $\beta_C,\mu$; learning rate $\gamma_t$; weight decay $\lambda$; scale constant $\tau$
\State Initialize $\mM=0$
\For{each optimizer step $t=0,1,\ldots$}
    \If{$t\bmod T=0$}
        \State Capture input activations $\mX_t^{\mathrm{cap}}$ and update each $\mC_{t,b}$ by Equation~\ref{eq:block-moment}
        \State Refresh each block basis $\mU_b$ from $\mC_{t,b}$ \Comment{Algorithm~\ref{alg:eig}}
    \EndIf
    \State $\mM\gets\mu\mM+\nabla_{\mW}f_{\mathcal{B}_t}(\mW)$
    \For{$b=1,\ldots,B$}
        \State $\mO_{:,\,\mathcal{S}_b}\gets
        \big(\operatorname{sign}(\mM_{:,\,\mathcal{S}_b}\mU_b)/\sqrt{m}\big)\mU_b^\top$
    \EndFor
    \State $\mW\gets(1-\gamma_t\lambda)\mW-\gamma_t\tau\sqrt{\max(m,n)}\,\mO$
\EndFor
\end{algorithmic}
\end{algorithm}

\paragraph{Update orthogonality under block partitioning.}
We compare how block partitioning and rank truncation change the singular-value spread of updates formed from the same momentum and second-moment state.
Table~\ref{tab:update-orthogonality} reports these updates at one GPT-2 Small checkpoint.
For the orthogonality metric defined in Table~\ref{tab:synthetic-orthogonality}, a rank-$r$ update is bounded above by $\sqrt{r/\min(m,n)}$.
At this checkpoint, increasing $B$ from 1 to 16 reduces \dobb's mean orthogonality from 0.849 to 0.815, whereas reducing Dion's rank from 768 to 48 reduces its orthogonality from 0.804 to 0.208.
This metric describes update geometry. 
Section~\ref{sec:results} separately reports validation loss under block partitioning.
\begin{table}[t]
\centering
\small
\caption{Update orthogonality under block partitioning and rank truncation.
Orthogonality is the metric defined in Table~\ref{tab:synthetic-orthogonality}.
Values average 72 weight matrices using the same momentum from a GPT-2 Small checkpoint at optimizer step 5{,}000 with second-moment capture at every step.
}
\label{tab:update-orthogonality}
\begin{tabular}{rrrr}
\toprule
\dobb blocks $B$ & \dobb orthogonality & Dion rank $r$ & Dion orthogonality \\
\midrule
1 & 0.849 & 768 & 0.804 \\
4 & 0.834 & 192 & 0.427 \\
16 & 0.815 & 48 & 0.208 \\
\bottomrule
\end{tabular}
\end{table}

\paragraph{Update magnitude and weight decay.}
Define $\operatorname{RMS}(\mO)=\|\mO\|_F/\sqrt{mn}$ before learning-rate multiplication and weight decay.
For complete orthonormal block bases and nonzero coefficient entries, \dobb's raw update has squared Frobenius norm $n$, independent of $B$, so its RMS is $1/\sqrt{m}$.
With learning rate $\gamma_t$, decoupled weight decay $\lambda$, and scale constant $\tau=0.2$, \dobb updates the weights by
\begin{equation}
\label{eq:dobb-scale}
\mW_{t+1}=(1-\gamma_t\lambda)\mW_t-\gamma_t s\,\mO_t,
\qquad
s=\tau\sqrt{\max(m,n)}.
\end{equation}
The update scale $s$ depends on the full matrix shape, not the local shard shape, and introduces no norm collective.
The update $s\mO_t$ therefore has RMS $\tau$ when $m\ge n$ and $\tau\sqrt{n/m}$ when $n>m$.
Appendix~\ref{app:update-rms} derives the update RMS of each optimizer and states the effects of numerical orthogonality and zero coefficients.

\paragraph{Locality.}
When each block lies within one shard, projection, element-wise sign normalization, and reconstruction use local operands.
Section~\ref{sec:comm} derives the resulting cross-shard payload and accounts separately for data-parallel second-moment synchronization.
\section{Communication and computation}
\label{sec:comm}

We count optimizer-specific cross-shard communication separately from the synchronization of second moments across data-parallel ranks.
All communication counts in this section, including Table~\ref{tab:comm-shard}, describe logical payloads.
We assume the Megatron-LM layout of Section~\ref{sec:background} under TP and, under either layout, that each block of \dobb lies within one shard.
When neither axis of a matrix is sharded, as under data parallelism alone, cross-shard communication is zero for every optimizer.

\paragraph{Blocks within shards.}
For \dobb on a row-parallel layer, or under FSDP with the input-axis sharding of Section~\ref{sec:background}, each input block and its momentum columns reside on the same rank.
On a column-parallel layer the input is replicated, so each rank can form every input-side block basis and apply it to its own momentum rows.
The $m$ in the sign divisor is known from the parameter shape.
No cross-shard statistic is needed for either update.
Thus, when each block lies within one shard, \dobb introduces zero optimizer-specific cross-shard bytes, including basis construction, beyond the second-moment synchronization across data-parallel ranks described below.

\begin{table}[t]
\centering
\small
\caption{Leading optimizer-specific cross-shard payload for one $m\times n$ matrix.
The counts apply when either the input or output axis is sharded, under TP or FSDP.
For \dobb, each block lies within one shard.
Constant collective factors and element sizes are omitted.
The counts exclude the data-parallel synchronization of \dobb's block second moments, $n^2/B$ entries per matrix at each refresh.
}
\label{tab:comm-shard}
\begin{tabular}{ll}
\toprule
Method & Optimizer-specific cross-shard payload \\
\midrule
Muon, dense exchange & $\mathcal{O}(mn)$ \\
Dion, rank $r$ & $\mathcal{O}(r(m+n))$ \\
\dobb & $0$ \\
\bottomrule
\end{tabular}
\end{table}

For a square $4096\times4096$ attention projection of Qwen3-8B in fp32, one dense payload is 67.11~MB, and two factors at $r=64$ contain 2.10~MB.
With each block within one shard, sign normalization contributes 0~MB of cross-shard payload.

\paragraph{FSDP has a separate synchronization requirement.}
FSDP ranks process different data, so consistent second moments require aggregation across the data-parallel group even when each block lies within one shard.
The ranks of a TP group process the same data, so this aggregation does not cross TP ranks.
\dobb captures and synchronizes its block second moments only on refresh iterations, once every $T$ optimizer steps; the other optimizer steps issue no optimizer collective.
Our implementation reduce-scatters the block second moments across the FSDP ranks, followed by an all-reduce across data-parallel replicas when the FSDP group is replicated.
Each rank then refreshes the bases of its own blocks, so no bases are exchanged.
Each refresh synchronizes $n^2/B$ second-moment entries per matrix (Appendix~\ref{app:costs}), separately from the cross-shard count above.

\paragraph{Arithmetic and memory.}
For \dobb, projection and reconstruction cost $\mathcal{O}(mn^2/B)$, while one orthogonal iteration per refresh costs $\mathcal{O}(n^3/(B^2T))$ amortized.
Capturing $N_c$ token positions once per refresh adds $\mathcal{O}(N_c n^2/(BT))$.
Storing the second moment and full block basis requires $2n^2/B$ scalars, in addition to momentum.
Small products, basis orthogonalization, and communication overlap determine the realized cost.

\paragraph{Comparison with a rank budget.}
We pair $B$ with Dion rank fraction $1/B$ in the quality comparison.
For a square $n\times n$ matrix with $r=n/B$, \dobb's block basis and Dion's persistent factor each contain $n^2/B$ entries.
Including momentum and the block second moment, \dobb stores $n^2+2n^2/B$ scalars, while Dion stores $n^2+n^2/B$.
Appendix~\ref{app:costs} gives the rectangular-matrix counts and distinguishes persistent state from transient workspace.

\section{Experiments}
\label{sec:setup}

We evaluate whether block-diagonal second moments retain optimization quality as their distributed cost decreases.
The principal quality comparison covers GPT-2 Small, Medium, Large, and XL \citep{radford2019gpt2}, with 124M, 354M, 772M, and 1.56B parameters excluding position embeddings.
The quality runs use OpenWebText \citep{gokaslan2019openwebtext}, sequence length 1024, a global batch of 480 sequences, and bfloat16 computation.
Table~\ref{tab:lr-sweep} records the scale-specific training budgets, and Table~\ref{tab:protocol} records the selected learning rates.
All configurations within a scale use the same token budget, data order, evaluation procedure, and learning-rate schedule.

\paragraph{Baselines and parameter groups.}
We compare \dobb at $B\in\{1,4,16,64\}$ against AdamW, Muon, RMNP, and Dion at rank fractions $1,1/4,1/16,1/64$ \citep{loshchilov2019adamw,jordan2024muon,deng2026rmnp,ahn2025dion}.
A Dion rank fraction $\rho$ specifies $r=\max(1,\lfloor\rho\min(m,n)\rfloor)$ for each weight matrix.
Matrix optimizers update the two-dimensional weights inside transformer blocks; embeddings, the language-model head, biases, and normalization parameters use AdamW.
Fused QKV parameters are treated as three weight matrices for matrix-update normalization, with each matrix's own $m,n$.
The quality implementation follows the RMNP/nanoGPT training framework \citep{deng2026rmnp,karpathy2022nanogpt}.

\paragraph{A shared update-scale protocol.}
Following \citet{liu2025muonscalable}, Muon, Dion, RMNP, and \dobb multiply their raw updates by the update scale $s=\tau\sqrt{\max(m,n)}$ of Equation~\ref{eq:dobb-scale}, with $\tau=0.2$, before learning-rate multiplication and weight decay.
The purpose of this scale is to bring the update RMS into the range of 0.2 to 0.4 that \citet{liu2025muonscalable} observe for AdamW, so that the matrix optimizers can share AdamW's learning rate and weight decay.
For a full-rank orthogonalized $m\times n$ update, their Lemma~1 gives raw RMS $1/\sqrt{\max(m,n)}$, so this scale gives Muon an update RMS of 0.2.
For the other optimizers, the update RMS depends on the matrix shape and, for Dion, on the rank (Appendix~\ref{app:update-rms}).
In our GPT-2 models, the update RMS of RMNP and \dobb is 0.2 or 0.4, also within this range.
Dion's update RMS is $0.2\sqrt{\rho}$, which falls from 0.2 at full rank to 0.025 at $\rho=1/64$ and lies below this range at every $\rho<1$; Appendix~\ref{app:update-rms} tests a rank-corrected scale that holds it at 0.2.
We select AdamW's learning rate by grid search at Small and Medium and extrapolate a power-law fit of the sweeps to Large and XL (Appendix~\ref{app:lrfit}), then transfer the selected rate and schedule to all other optimizers.
All optimizers use weight decay 0.1, following the pretraining setting of \citet{liu2025muonscalable}.

\subsection{Quality under block and rank compression}
\label{sec:results}

Table~\ref{tab:quality} reports final validation loss under the shared update-scale protocol.
Varying \dobb's block count and Dion's rank fraction measures the quality cost of block partitioning and rank restriction, respectively.

\begin{table}[t]
\centering
\small
\caption{Final validation loss under the shared update-scale protocol, with common learning rate and weight decay within each model scale.
Small, Medium, Large, and XL train for 10{,}000, 20{,}000, 30{,}000, and 40{,}000 steps.
Dion ranks are fractions of each weight matrix's full rank.
}
\label{tab:quality}
\begin{threeparttable}
\begin{tabular}{llrrrr}
\toprule
Optimizer & Blocks / rank fraction & Small & Medium & Large & XL \\
\midrule
AdamW & n/a & 3.044 & 2.761 & 2.606 & 2.497 \\
Muon & full & 3.004 & 2.725 & 2.574 & 2.460 \\
RMNP & n/a & 3.083 & 2.778 & 2.618 & 2.499 \\
\midrule
Dion & $1$ & 3.002 & 2.720 & 2.568 & 2.453 \\
 & $1/4$ & 3.038 & 2.754 & 2.596 & 2.481 \\
 & $1/16$ & 3.108 & 2.802 & 2.636 & 2.511\rlap{$^{\dagger}$} \\
 & $1/64$ & 3.265 & 2.879 & 2.696 & 2.558 \\
\midrule
\dobb & $B=1$ & 3.017 & 2.735 & 2.579 & 2.461 \\
 & $B=4$ & 3.024 & 2.741 & 2.588 & 2.471 \\
 & $B=16$ & 3.026 & 2.746 & 2.595 & 2.476\rlap{$^{\dagger}$} \\
 & $B=64$ & 3.037 & 2.742 & 2.600 & 2.482 \\
\bottomrule
\end{tabular}
\begin{tablenotes}[flushleft]
\footnotesize
\item[$^\dagger$] The abstract and Section~\ref{sec:results-fsdp} compare this GPT-2 XL pair, whose optimizer-call and pooled iteration times appear in Table~\ref{tab:gpt2-h100}.
\end{tablenotes}
\end{threeparttable}
\end{table}

From $B=1$ to $B=64$, \dobb's validation loss rises from 3.017 to 3.037 on Small, from 2.735 to 2.742 on Medium, from 2.579 to 2.600 on Large, and from 2.461 to 2.482 on XL, increases of 0.020, 0.007, 0.021, and 0.021.
Reducing Dion's rank fraction from 1 to $1/64$ raises its loss from 3.002 to 3.265, from 2.720 to 2.879, from 2.568 to 2.696, and from 2.453 to 2.558, increases of 0.263, 0.159, 0.128, and 0.105.
At $B=1$, \dobb has higher loss than Muon and full-rank Dion at each scale; at $B\in\{4,16,64\}$, it has lower loss than Dion at the corresponding rank fraction $1/B$.

\subsection{Iteration time under FSDP and tensor parallelism}
\label{sec:results-fsdp}

\begin{table}[t]
\centering
\footnotesize
\setlength{\tabcolsep}{1.85pt}
\caption{Times in milliseconds, with $\DP$ data-parallel replicas and FSDP shard degree $\FS$.
GPT-2 3B, 6B, and 8B are scaled-up GPT-2-style models.
Optimizer call is the mean duration of \texttt{optimizer.step()}; \dobb's is averaged over one refresh cycle, $(\text{Refresh}+8\cdot\text{Non-refresh})/9$.
Iteration is the pooled iteration time defined in Appendix~\ref{app:pooled}.
Dion uses rank $r=400$ on Large; rank fraction $\rho=1/3$ on XL (A5000), 6B, and 8B; $\rho=1/4$ on XL (H100) and 3B; and on Qwen3, a rank fraction matched to the block width $1024/\FS$ (Appendix~\ref{app:distributed-qwen}).
For GPT-2 XL on H100, Table~\ref{tab:gpt2-h100} also reports Dion at $\rho=1/16$, the rank fraction of the loss comparison in the abstract, with an optimizer call of 124.1~ms.
Dashes mark methods not run on Qwen3.
Appendix~\ref{app:distributed} gives \dobb's \texttt{step()} times on refresh and non-refresh iterations and all other configurations.
}
\label{tab:fsdp}
\begin{tabular}{@{}llrrrrrrrrrrrr@{}}
\toprule
 & & & & & & \multicolumn{4}{c}{Optimizer call (ms)} & \multicolumn{4}{c}{Iteration (ms)} \\
\cmidrule(lr){7-10}\cmidrule(lr){11-14}
Model & GPU & $\DP$ & $\FS$ & $\TP$ & GPUs & AdamW & Muon & Dion & \dobb & AdamW & Muon & Dion & \dobb \\
\midrule
GPT-2 Large & A5000 & 1 & 10 & 2 & 20 & 11.3 & 623.0 & 1224.1 & 6.9 & 7764 & 8376 & 8977 & 7760 \\
GPT-2 XL & A5000 & 2 & 8 & 2 & 32 & 18.7 & 946.5 & 2523.2 & 16.3 & 11,649 & 12,577 & 14,153 & 11,646 \\
GPT-2 XL & H100 & 2 & 16 & 1 & 32 & 9.3 & 82.7 & 232.7 & 9.2 & 1069 & 1143 & 1293 & 1069 \\
GPT-2 3B & H100 & 1 & 16 & 2 & 32 & 49.4 & 454.3 & 1385.6 & 7.6 & 2388 & 2793 & 3724 & 2346 \\
GPT-2 6B & H100 & 1 & 8 & 4 & 32 & 13.9 & 2106.8 & 1505.7 & 10.9 & 4823 & 6916 & 6315 & 4820 \\
GPT-2 8B & H100 & 1 & 16 & 4 & 64 & 15.2 & 2782.1 & 2081.5 & 10.0 & 5337 & 8104 & 7403 & 5332 \\
\midrule
Qwen3-8B & H100 & 1 & 8 & 1 & 8 & -- & -- & 159.5 & 86.2 & -- & -- & 423 & 349 \\
Qwen3-32B & H100 & 1 & 32 & 2 & 64 & -- & -- & 1307.7 & 168.0 & -- & -- & 17,664 & 16,524 \\
\bottomrule
\end{tabular}
\end{table}

We measured iteration time on GPT-2-style models from 772M to 8.1B parameters with our FSDP and TP implementation, and on Qwen3-8B and Qwen3-32B \citep{yang2025qwen3} with an implementation built on the open-source Dion codebase \citep{ahn2025dion}.
Both implementations use PyTorch FSDP2 with DTensor-based TP \citep{liang2025torchtitan}.
In all timing runs, NCCL communicated between nodes over TCP sockets.
Table~\ref{tab:fsdp} reports one configuration per model and GPU type; Appendix~\ref{app:distributed} gives the setups and the remaining configurations.
Because \dobb's forward-backward overhead lies within the run-to-run variation of forward-backward time, the table reports pooled iteration times, which share one forward-backward estimate per configuration and add each method's measured optimizer call (Appendix~\ref{app:pooled}).
In Table~\ref{tab:fsdp}, \dobb's amortized optimizer call takes 6.9 to 16.3~ms on GPT-2 and 86.2 to 168.0~ms on Qwen3, against 82.7 to 2782.1~ms for Muon and 159.5 to 2523.2~ms for Dion.
Muon's pooled iteration time is 1.07 to 1.52 times that of \dobb, and Dion's is 1.07 to 1.59 times.
The quality runs and the timing runs both include GPT-2 XL with $B=16$ for \dobb and Dion at the corresponding rank fraction $1/16$.
In this configuration, \dobb reaches a validation loss of 2.476 against 2.511 for Dion (Table~\ref{tab:quality}), and under FSDP on 32 H100 GPUs, its amortized optimizer call takes 9.2~ms against 124.1~ms, with pooled iteration times of 1069.1 and 1184.0~ms (Table~\ref{tab:gpt2-h100}).

\section{Conclusion}
\label{sec:conclusion}

We introduce \dobb, which normalizes momentum coefficients in block-diagonal activation second-moment eigenbases.
The central design choice is to discard cross-block correlations while retaining every basis direction within each block.
Together with element-wise sign normalization, this construction introduces zero optimizer-specific cross-shard bytes when each block lies within one shard, beyond the data-parallel synchronization of the block second moments.

Across GPT-2 Small, Medium, Large, and XL, \dobb exhibits less validation-loss degradation under block partitioning than Dion under corresponding rank reductions.
\dobb's optimizer call is shorter than those of Muon and Dion in every measured FSDP and TP configuration (Appendix~\ref{app:distributed}).
These timing runs are short and report pooled iteration times.

Further work could investigate how block size and basis-refresh frequency should vary across layers and training stages.
Because \dobb's basis refresh reads no gradient, it could also move out of the optimizer call into idle intervals of forward-backward computation and communication (Appendix~\ref{app:off-step}).
Understanding how discarded cross-block correlations and coefficient normalization theoretically affect optimization could also guide block partitions that preserve useful structure while respecting shard boundaries.

\clearpage
\bibliography{references}
\bibliographystyle{plainnat}

\clearpage
\appendix
\etocdepthtag.toc{appendix}
\etocsettagdepth{main}{none}
\etocsettagdepth{appendix}{subsection}
\etocsettocstyle{\section*{Appendix contents}}{}
\tableofcontents
\clearpage

\section{Notation}
\label{app:notation}

Table~\ref{tab:notation} lists the symbols that recur across sections, grouped by topic, with the location of each definition.
Bold capitals denote matrices, and bold lowercase letters denote vectors.
Calligraphic letters denote index sets, the mini-batch, the coefficient normalization, the fitted loss, and asymptotic order $\mathcal{O}(\cdot)$.
For a matrix $\mM$, $\mM_{:,\mathcal{S}}$ selects the columns indexed by $\mathcal{S}$ and $\mM_{\mathcal{S},:}$ the rows.
Matrix dimensions refer to the full weight matrix after the QKV split and before sharding.

\begin{table}[!ht]
\centering
\footnotesize
\caption{Notation.
Parenthesized values are the principal settings.
}
\label{tab:notation}
\begin{tabular}{@{}l p{0.61\linewidth} l@{}}
\toprule
Symbol & Meaning & Defined in \\
\midrule
\multicolumn{3}{@{}l}{\emph{Layer and gradient statistics}} \\
$m$, $n$ & Output and input dimensions of a weight matrix & Section~\ref{sec:background} \\
$\mW_t$, $t$ & Weight matrix of a layer $\vy=\mW\vx$ at optimizer step $t$ & Section~\ref{sec:background} \\
$\mathcal{B}_t$, $N$ & Mini-batch at step $t$, and its number of examples or token positions & Section~\ref{sec:background} \\
$f_{\mathcal{B}_t}$ & Mean loss on $\mathcal{B}_t$ & Equation~\ref{eq:batch-gradient} \\
$\vx_i$, $\vg_i$ & Input activation and output gradient of the $i$th contribution & Equation~\ref{eq:batch-gradient} \\
$\mX_t$, $\mD_t$ & Stacked activations $[\vx_1,\ldots,\vx_N]$ and output gradients $[\vg_1,\ldots,\vg_N]$; the superscript $\mathrm{cap}$ restricts these matrices to captured positions & Equation~\ref{eq:batch-gradient} \\
$\mV_X$, $\mSigma_X$ & Left singular vectors and diagonal singular-value matrix in the compact SVD $\mX_t^\top=\mV_X\mSigma_X\mU^\top$ & Section~\ref{sec:background} \\
$\mA_t$, $\mG_t$ & Activation and output-gradient second moments, $n\times n$ and $m\times m$ & Equation~\ref{eq:factors} \\
\midrule
\multicolumn{3}{@{}l}{\emph{Blocks and bases}} \\
$B$, $b$ & Number of equal-sized blocks per weight matrix, and block index & Section~\ref{sec:background} \\
$\mathcal{S}_b$ & Coordinates of block $b$ on the input axis for $A$, output axis for $G$ & Equation~\ref{eq:block-moment} \\
$\mS_{t,b}$ & Block second moment of one capture window & Equation~\ref{eq:block-moment} \\
$\mC_{t,b}$, $\beta_C$ & Exponentially averaged block second moment, and its decay (0.95) & Equation~\ref{eq:block-moment} \\
$T$ & Capture and basis-refresh period in optimizer steps (9) & Section~\ref{sec:method} \\
$N_c$ & Token positions captured per refresh & Equation~\ref{eq:block-moment} \\
$\varepsilon$ & Shift added to a block second moment: $10^{-6}$ for the cold-start eigendecomposition and Householder QR; $\omega\max_i\sum_j|C_{ij}|$ for Cholesky QR & Algorithm~\ref{alg:eig} \\
$\omega$ & Relative shift for Cholesky QR, $\varepsilon\ge\omega\lambda_{\max}$ ($10^{-2}$) & Algorithm~\ref{alg:eig} \\
$\mU$, $\mU_{t,b}$ & Orthonormal basis in the update template, and orthonormal block basis in use at step $t$, approximating the eigenbasis of $\mC_{t,b}$ & Equation~\ref{eq:template} \\
\midrule
\multicolumn{3}{@{}l}{\emph{Update}} \\
$\mM$, $\mM_t$ & Momentum & Equation~\ref{eq:template} \\
$\mu$ & Momentum decay (0.95) & Algorithm~\ref{alg:dobb} \\
$\mZ_{t,b}$ & Coefficients of the momentum slice in the block basis & Equation~\ref{eq:dobb-update} \\
$\mathcal{N}$, $\widehat{\mZ}_b$ & Coefficient normalization, and normalized coefficients $\widehat{\mZ}_b=\mathcal{N}(\mZ_b)$ & Equation~\ref{eq:template} \\
$\ell_\infty$, $\ell_2$ & Sign normalization $\operatorname{sign}(\mZ)/\sqrt{d}$, and unit-norm columns & Algorithm~\ref{alg:dobb-full} \\
$d$ & Length of the non-basis axis: $m$ for $A$, $n$ for $G$ & Algorithm~\ref{alg:dobb-full} \\
$\mO$, $\mO_t$ & Output of the update template, and raw update assembled from the blocks, before scaling & Equation~\ref{eq:template} \\
$\operatorname{RMS}(\mO)$ & Per-element root mean square, $\|\mO\|_F/\sqrt{mn}$ & Section~\ref{sec:method} \\
$\tau$, $s$ & Scale constant (0.2), and update scale $s=\tau\sqrt{\max(m,n)}$ & Equation~\ref{eq:dobb-scale} \\
$\gamma_t$, $\lambda$ & Learning rate at step $t$, and decoupled weight decay (0.1) & Equation~\ref{eq:dobb-scale} \\
\midrule
\multicolumn{3}{@{}l}{\emph{Baselines and parallelism}} \\
$r$ & Dion rank & Section~\ref{sec:background} \\
$\rho$ & Dion rank fraction, with $r=\max(1,\lfloor\rho\min(m,n)\rfloor)$ & Section~\ref{sec:setup} \\
$\TP$ & Tensor-parallel degree & Section~\ref{sec:background} \\
$\FS$ & FSDP shard degree & Section~\ref{sec:background} \\
$\DP$ & Data-parallel replication degree: the number of model replicas, each sharded over $\FS\cdot\TP$ ranks & Table~\ref{tab:fsdp} \\
\bottomrule
\end{tabular}
\end{table}

\section{Complete algorithm and implementation settings}
\label{app:algorithm}

Algorithm~\ref{alg:dobb-full} specifies the four Dobb variants with the update scale of Equation~\ref{eq:dobb-scale}.
Algorithm~\ref{alg:eig} maintains a complete orthonormal basis within each block.
The update uses only the eigenvectors of each block second moment: the sign and the column $\ell_2$ normalization both ignore positive scaling of coefficient columns, so an inverse-eigenvalue rescaling of the coefficients would not change the update.
Each warm-started refresh orthonormalizes $(\mC+\varepsilon\mI)\mU$, and we use two QR routines for this step.
Householder QR uses the fixed shift $\varepsilon$ of the cold-start eigendecomposition, and Cholesky QR sets $\varepsilon=\omega\max_i\sum_j|C_{ij}|$ instead.
Cholesky QR replaces the sequence of Householder reflections with matrix products, a Cholesky factorization of the Gram matrix $\mQ^\top\mQ$, and a triangular solve.
The Gram matrix has the squared condition number of $\mQ$, so its factorization can fail in fp32 when $\mQ$ is ill-conditioned.
For an orthonormal $\mU$, the condition number of $(\mC+\varepsilon\mI)\mU$ equals that of $\mC+\varepsilon\mI$, which the shift $\varepsilon\ge\omega\lambda_{\max}(\mC)$ bounds by $(1+\omega)/\omega$, or 101 at $\omega=10^{-2}$.
The shift leaves the eigenvectors unchanged but replaces each eigenvalue ratio $\lambda_i/\lambda_j$ by $(\lambda_i+\varepsilon)/(\lambda_j+\varepsilon)$, which slows the convergence of orthogonal iteration.
A distributed implementation may assign basis factorizations to different ranks, batch matrices with equal block shapes, and distribute refreshed bases to the ranks that consume them.
Those assignments change which collectives run and when, not the update that each block receives.

\begin{algorithm}[h]
\caption{Full block-basis refresh.
The first call uses an exact eigendecomposition; subsequent calls apply one step of warm-started orthogonal iteration, orthonormalized by Householder QR or by Cholesky QR (Algorithm~\ref{alg:cholqr}).
Equal-sized blocks can be processed in a batch.}
\label{alg:eig}
\begin{algorithmic}[1]
\Function{Eigenvectors}{$\mC,\mU,\varepsilon,\omega$}
    \If{$\mU$ is empty}
        \State $\mU\gets$ eigenvectors from $\textproc{Eigh}(\mC+\varepsilon\mI)$
        \State Discard the eigenvalues and retain every column of $\mU$
    \ElsIf{the routine is Householder QR}
        \State $\mU\gets$ orthonormal factor of $\textproc{HouseholderQR}\big((\mC+\varepsilon\mI)\mU\big)$
    \Else
        \State $\varepsilon\gets\omega\max_i\sum_j|C_{ij}|$ \Comment{upper bound on $\omega\lambda_{\max}(\mC)$}
        \State $\mU\gets\textproc{CholeskyQR}\big((\mC+\varepsilon\mI)\mU\big)$ \Comment{Algorithm~\ref{alg:cholqr}}
    \EndIf
    \State \Return $\mU$
\EndFunction
\end{algorithmic}
\end{algorithm}

\begin{algorithm}[ht]
\small
\caption{Complete update for the four Dobb variants on one matrix of a single layer with the update scale of Equation~\ref{eq:dobb-scale}.
The side and norm select the variant: $(A,\ell_\infty)$ gives \dobb, $(G,\ell_\infty)$ \dobbG, $(A,\ell_2)$ \dobbLtwo, and $(G,\ell_2)$ \dobbGLtwo.
Appendices~\ref{app:norm} and~\ref{app:basis} compare the four variants.
The gradient uses the full mini-batch $\mathcal{B}_t$; capture may use a subset of its micro-batches.
The per-matrix computation is shown; synchronization follows the actual TP/FSDP mesh. The optimizer-step index $t$ is omitted from the momentum, bases, coefficients, and update in both this algorithm and Algorithm~\ref{alg:dobb}.}
\label{alg:dobb-full}
\begin{algorithmic}[1]
\Require $\mW\in\mathbb{R}^{m\times n}$; side $A$ or $G$ (basis from $\mA_t$ or $\mG_t$); norm $\ell_\infty$ or $\ell_2$; blocks $\{\mathcal{S}_b\}_{b=1}^{B}$ of the input coordinates for side $A$, of the output coordinates for side $G$
\Require $\beta_C,\mu,T,\varepsilon,\omega,\tau,\lambda,\gamma_t$; capture policy; QR routine
\State $d\gets m$ for side $A$, otherwise $n$
\State Initialize $\mM=0$; block statistics and bases are empty
\For{each optimizer step $t=0,1,\ldots$}
    \If{$t\bmod T=0$}
        \State Capture $\mX_t^{\mathrm{cap}}$ for side $A$ or $\mD_t^{\mathrm{cap}}$ for side $G$ from the micro-batches selected by the capture policy
        \State $\mS_{t,b}\gets$ mean second moment of the captured positions for each block $b$, as in Equation~\ref{eq:block-moment}
        \State Synchronize $\mS_{t,b}$ across data-parallel ranks when required
        \State $\mC_{t,b}\gets\mS_{t,b}$ on the first capture, otherwise $\beta_C\mC_{\mathrm{prev},b}+\mS_{t,b}$
        \State $\mU_b\gets\textproc{Eigenvectors}(\mC_{t,b},\mU_b,\varepsilon,\omega)$ for each block $b$
    \EndIf
    \State $\mM\gets\mu\mM+\nabla_{\mW}f_{\mathcal{B}_t}(\mW)$
    \For{each locally needed block $b$}
        \State $\mV_b\gets\mM_{:,\,\mathcal{S}_b}$ for side $A$, otherwise $\mM_{\mathcal{S}_b,:}^\top$
        \State $\mZ_b\gets\mV_b\mU_b$
        \If{norm is $\ell_\infty$}
            \State $\widehat{\mZ}_b\gets\operatorname{sign}(\mZ_b)/\sqrt{d}$
        \Else
            \State $\widehat{\mZ}_b\gets\mZ_b$ with each nonzero column scaled to unit $\ell_2$ norm; zero columns stay zero
        \EndIf
        \State $\widehat{\mV}_b\gets\widehat{\mZ}_b\mU_b^\top$
        \State Store $\widehat{\mV}_b$ in $\mO_{:,\,\mathcal{S}_b}$ for side $A$, otherwise its transpose in $\mO_{\mathcal{S}_b,:}$
    \EndFor
    \State $\mW\gets(1-\gamma_t\lambda)\mW-\gamma_t\tau\sqrt{\max(m,n)}\,\mO$
\EndFor
\end{algorithmic}
\end{algorithm}

\begin{algorithm}[ht]
\caption{Cholesky QR with two passes.
Algorithm~\ref{alg:eig} applies it to $(\mC+\varepsilon\mI)\mU$ with $\varepsilon\ge\omega\lambda_{\max}(\mC)$, which bounds the condition number of the input by $(1+\omega)/\omega$.}
\label{alg:cholqr}
\begin{algorithmic}[1]
\Function{CholeskyQR}{$\mQ$}
    \For{pass $=1,2$}
        \State $\mR\gets\textproc{Cholesky}(\mQ^\top\mQ)$ \Comment{upper triangular, $\mQ^\top\mQ=\mR^\top\mR$}
        \State $\mQ\gets\mQ\mR^{-1}$ \Comment{triangular solve}
    \EndFor
    \State \Return $\mQ$
\EndFunction
\end{algorithmic}
\end{algorithm}

Each block has width $n/B$ on the activation side and $m/B$ on the output side; every block count in our experiments divides the length of the partitioned axis.
A local update requires only that each block lie within one shard.
The block count is defined per weight matrix; a fused QKV parameter is split into its three weight matrices before the update, and each receives its own blocks.
No equality of block width across differently shaped layers is assumed.

The principal capture policy uses only the first micro-batch on refresh iterations.
The matrix $\mX_t^{\mathrm{cap}}$ contains the $N_c$ captured input columns, and $\mD_t^{\mathrm{cap}}\in\mathbb{R}^{m\times N_c}$ contains the corresponding output-gradient columns for side $G$.
With all-micro-batch capture, form a token-weighted second moment across the captured micro-batches before applying the window-level decay.
This normalization makes the second moment a mean over captured tokens for either capture policy in Appendix~\ref{app:micro0}.

Second-moment capture uses forward hooks for side $A$ and backward hooks for side $G$.
The hooks are inactive outside training mode and on non-refresh iterations, and they accumulate the block outer products in place, so captured activations or output gradients are not retained after the hook returns.
Column normalization in the $\ell_2$ variants sums squares over a full coefficient column; when that axis is sharded, the sum requires one reduction across the shards, which Section~\ref{sec:comm} counts separately from the local update.
The principal settings are $\beta_C=0.95$, $\mu=0.95$, $T=9$, one warm orthogonal iteration, fp32 second-moment accumulation, and bf16 basis storage.
Appendix~\ref{app:precision} compares these two precision choices against their alternatives.
Appendix~\ref{app:orthiter} compares the single orthogonal iteration with two and three iterations and with an exact eigendecomposition.
The shift for the cold-start eigendecomposition and Householder QR is $\varepsilon=10^{-6}$ in the stated accumulation convention.
The scale constant is $\tau=0.2$, and weight decay is $\lambda=0.1$.
Table~\ref{tab:protocol} lists the selected learning rates.

\section{Communication, arithmetic, and state accounting}
\label{app:costs}

\paragraph{Communication for the ablation configurations.}
Algorithm~\ref{alg:dobb-full} specifies three alternative configurations that change the basis source or coefficient normalization of \dobb, and Appendix~\ref{app:ablations} compares them.
With each block within one shard, sign normalization requires no cross-shard reduction for either basis source.
Column normalization changes this result only when its reduction axis is sharded.
\dobbLtwo reduces one squared norm for each of $n$ coefficient columns when the output axis is sharded.
\dobbGLtwo reduces $m$ such values when the input axis is sharded.
Independent norm reductions can be packed to reduce collective-launch overhead.
For a square $4096\times4096$ matrix in fp32, these 4096 scalars contain 0.0164~MB.
\begin{table}[ht]
\centering
\small
\caption{Leading optimizer-specific cross-shard payload for the basis and normalization ablations on one $m\times n$ matrix, with each block within one shard.
The $\ell_2$ entries are scalars reduced per update; constant collective factors and element sizes are omitted.
The counts exclude the data-parallel synchronization of the block second moments.}
\label{tab:comm-ablations}
\begin{tabular}{lll}
\toprule
Configuration & Output axis $m$ sharded & Input axis $n$ sharded \\
\midrule
\dobb & $0$ & $0$ \\
\dobbG & $0$ & $0$ \\
\dobbLtwo & $\mathcal{O}(n)$ & $0$ \\
\dobbGLtwo & $0$ & $\mathcal{O}(m)$ \\
\bottomrule
\end{tabular}
\end{table}

\paragraph{Object size and wire traffic.}
For input-axis blocks of width $n/B$, $\sum_b|\mathcal{S}_b|^2=n^2/B$.
For output-axis blocks the corresponding sum is $m^2/B$.
Multiplying by the bytes per stored scalar gives the second-moment object size.
Communicating this object once per $T$ updates gives an amortized size of $n^2/(BT)$ or $m^2/(BT)$ scalars.
For a $4096\times4096$ matrix at $B=16$, $T=9$, and fp32 accumulation, the second-moment object contains 4{,}194{,}304 bytes per refresh, or approximately 0.466~MB per optimizer step.
A basis exchange is a separate object whose size depends on its storage precision.
For a ring all-reduce over $R$ ranks, each rank sends $2(R-1)/R$ times the input-buffer size; reduce-scatter and all-gather have different accounting.

\paragraph{Projection and basis maintenance.}
For input block widths $n_b$, projection and reconstruction require work proportional to $m\sum_b n_b^2$.
For $B$ equal input blocks of width $n/B$, projection and reconstruction each cost $\mathcal{O}(B\,m(n/B)^2)=\mathcal{O}(mn^2/B)$.
Orthogonal iteration and QR require work proportional to $\sum_b n_b^3$, giving $n^3/B^2$ per iteration.
An exact cold-start eigendecomposition has the same cubic order and occurs once.
Captured outer products cost $N_c\sum_b n_b^2$.
These counts exclude data movement, launch overhead, and overlap with the forward-backward pass.
For output-side blocks of widths $m_b$, use $n\sum_b m_b^2$, $\sum_b m_b^3$, and $N_c\sum_b m_b^2$, respectively.

\paragraph{State versus workspace.}
The four Dobb variants retain momentum of size $mn$, a block second moment, and a block basis.
Their element counts are $mn+2n^2/B$ on the activation side and $mn+2m^2/B$ on the output side, before sharding and precision are applied.
Captured activations, projected coefficients, and factorization buffers are additional transient memory.
Dion retains momentum of size $mn$ and a factor of size $r\min(m,n)$; its leading factor-communication payload is $r(m+n)$.
Error feedback may be represented within an existing momentum buffer rather than a separate dense allocation.
Dion also uses a second factor and QR workspace during an update.
Storage precision and the device mesh determine the byte count per rank for each implementation.

\section{Construction of the synthetic matrices}
\label{app:synthetic}

Table~\ref{tab:synthetic-orthogonality} reports update orthogonality computed from synthetic matrices.
The GPT-2 Small second-moment spectra in Figure~\ref{fig:rank}, measured at 50\% of training, guide the spectral shapes and effective-rank levels in panel~(b).
All matrices used here are synthetic.
The synthetic gradients and the matrix $\mY$ are $768\times768$, and orthogonality is the metric defined in the caption of Table~\ref{tab:synthetic-orthogonality}.

\paragraph{Matrices with prescribed effective rank.}
In panel~(a), $\mE_Y$ and $\mP_Y$ are independent, uniformly distributed random orthogonal $768\times768$ matrices, and $\mY=\mE_Y\operatorname{diag}(1,2^{-\kappa_Y},\ldots,768^{-\kappa_Y})\,\mP_Y^\top$.
For each target, bisection on the exponent $\kappa_Y$ set the effective rank of $\mY$ to the target, which gave $\kappa_Y=0.491$, 0.738, 0.997, and 1.307 for 80\%, 50\%, 20\%, and 5\%.
This effective rank is that of $\mY$ itself: it follows Figure~\ref{fig:rank} with $p_i$ the $i$th normalized singular value of $\mY$ rather than an eigenvalue of a second moment.

\paragraph{Synthetic gradients.}
Panel~(b) uses the mini-batch gradient $\mD\mX^\top/N$ of Equation~\ref{eq:batch-gradient} with $m=n=768$ and $N=2048$ token positions, omitting the step index $t$.
We set $\mX=\sqrt{N}\,\mE_X\mLambda_X^{1/2} \mV_X^\top$ and $\mD=\sqrt{N}\,\mE_D\mLambda_D^{1/2}\mP_D^\top$, where $\mE_X$ and $\mE_D$ are random $768\times768$ orthogonal matrices, $ \mV_X$ and $\mP_D$ are random $N\times768$ matrices with orthonormal columns, and $\mLambda_X$ and $\mLambda_D$ are diagonal with prescribed positive entries in decreasing order.
The second moments of Equation~\ref{eq:factors} are then exactly $\mA=\mE_X\mLambda_X\mE_X^\top$ and $\mG=\mE_D\mLambda_D\mE_D^\top$, and the effective ranks in panel~(b) are those of $\mA$ and $\mG$, computed from their eigenvalues as in Figure~\ref{fig:rank}.
The four random matrices are independent and uniformly distributed; independent factors match the interpretation of the K-FAC approximation as statistical independence between products of activations and products of output gradients \citep[Section~3.1]{martens2015kfac}.
The diagonal entries follow the shapes of the measured spectra: in 9 of the 12 activation spectra recorded in full, from four weight-matrix types in the first, sixth, and last layers, the largest eigenvalue lies 2 to 30 times above a power law fitted to the other eigenvalues, whereas no output-gradient spectrum has a separated largest eigenvalue.
We therefore gave the first entry of $\mLambda_X$ a fixed share of its trace and made the $i$th entry, for $i\ge2$, proportional to $i^{-\kappa_X}$; the $i$th entry of $\mLambda_D$ is proportional to $i^{-\kappa_D}$ for every $i$.

\paragraph{Effective-rank levels.}
At 50\% of training, the median effective rank over the 72 weight matrices of GPT-2 Small is 13\% for the activation second moment and 55\% for the output-gradient second moment; over the 48 square matrices of the query, key, value, and attention output projections, the medians are 17\% and 38\%.
We fixed the activation effective rank at 15\%, between the two activation medians, with a share of 25\% and $\kappa_X=0.94$, found by bisection along a line between the approximate ends of the measured ranges of the share (9\% to 75\%) and the exponent (0.74 to 1.6).
For the output-gradient levels of 88\%, 80\%, 72\%, 55\%, 38\%, and 22\%, bisection on $\kappa_D$ gave 0.40, 0.49, 0.57, 0.70, 0.83, and 0.97.
These levels include both output-gradient medians and lie within the measured range of 0.2\% to 92\%; the steepest exponent, 0.97, slightly exceeds the range of 0.34 to 0.92 measured for the output-gradient spectra that follow a power law.

\paragraph{Updates.}
Both updates use the exact eigenbasis $\mU$ of $\mA$ as a single block ($B=1$), computed in double precision, and apply Equation~\ref{eq:dobb-update} to the gradient in place of the momentum.
\dobb takes the sign of the coefficients, and \dobbLtwo, defined in Algorithm~\ref{alg:dobb-full}, scales each coefficient column to unit $\ell_2$ norm; positive scalar factors such as $1/\sqrt{m}$ do not change the orthogonality.
Up to the order and signs of its columns, which affect neither update, $\mU=\mE_X$, so $\mX^\top= \mV_X\mSigma_X\mU^\top$ with $\mSigma_X=\sqrt{N}\mLambda_X^{1/2}$ is the decomposition of Section~\ref{sec:background}, and the coefficients $\mD\mX^\top\mU/N=\mD \mV_X\mLambda_X^{1/2}/\sqrt{N}$ depend on the activation spectrum only through a positive scaling of whole columns.
Neither normalization changes under such a scaling, so the activation level does not affect panel~(b).

\paragraph{Scope of the construction.}
The construction compares the orthogonality of individual updates on square synthetic gradients with independent factors and an exact activation eigenbasis.
Rectangular matrices, structured singular vectors, and momentum with averaged or reused bases can change the observed orthogonality; the prescribed spectra also omit departures from a single power law in the query, key, and value factors.

\section{Training protocol and update scale}
\label{app:protocol}

\paragraph{AdamW learning-rate selection.}
The principal comparison transfers the AdamW-selected learning rate to every optimizer at a given scale.
Table~\ref{tab:protocol} lists the selected rate, evaluation, and seeds at each scale, and Table~\ref{tab:lr-sweep} lists the grids and training budgets.
Selection uses the same validation procedure for every AdamW candidate; the final benchmark uses the selected rate without optimizer-specific retuning.
Appendix~\ref{app:lrfit} fits the optimal learning rate as a power law in model size and extrapolates it to Large and XL.

\begin{table}[ht]
\centering
\small
\caption{Training protocol for the principal quality comparison.
Table~\ref{tab:lr-sweep} gives the architecture, step count, training token budget, and AdamW grid at each size.
Grid indices $j$ refer to Equation~\ref{eq:lr-grid}.
Validation loss is computed on a fixed set of held-out sequences.
}
\label{tab:protocol}
\begin{tabular}{lrrrr}
\toprule
Field & Small & Medium & Large & XL \\
\midrule
Selected learning rate & $3.54\times10^{-3}$ & $3.54\times10^{-3}$ & $2.5\times10^{-3}$ & $2.5\times10^{-3}$ \\
Grid index $j$ & $-1$ & $-1$ & $-2$ & $-2$ \\
Evaluation interval (steps) & 1{,}000 & 1{,}000 & 1{,}000 & 1{,}000 \\
Evaluation sequences & 3{,}000 & 3{,}000 & 3{,}000 & 3{,}000 \\
Seeds & 5000 & 5000 & 5000 & 5000 \\
\bottomrule
\end{tabular}
\end{table}

\subsection{Update RMS of each optimizer}
\label{app:update-rms}

Under the shared update-scale protocol, every matrix optimizer in Table~\ref{tab:update-rms} multiplies its raw update $\mO$ by the update scale $s=\tau\sqrt{\max(m,n)}$ of Equation~\ref{eq:dobb-scale} before learning-rate multiplication and weight decay.
This scale replaces, rather than compounds, any earlier shape correction.
All matrix dimensions refer to the full weight matrix before sharding and after the declared QKV split.
Because the update scale is shared, the update RMS, $\operatorname{RMS}(s\mO)=s\|\mO\|_F/\sqrt{mn}$, differs across optimizers only through the Frobenius norm of the raw update.
\citet{liu2025muonscalable} introduced this scale to bring Muon's update RMS into the range of 0.2 to 0.4 that they observe for AdamW, and they report that Muon can then reuse the learning rate and weight decay tuned for AdamW.
In our GPT-2 models, the resulting update RMS of RMNP, the four Dobb variants, and full-rank Dion also lies between 0.2 and 0.4.
Under this shared update-scale protocol, we use the AdamW-selected learning rate and weight decay for all methods.

\begin{table}[ht]
\centering
\small
\caption{Raw-update norm, update scale $s$, and resulting update RMS of each matrix optimizer in exact arithmetic with nonzero coefficients.
Approximate orthogonalization and zero columns can lower or perturb the attained RMS.
The GPT-2 column uses $\tau=0.2$ and the rank fraction $\rho=r/\min(m,n)$.
Every entry of the GPT-2 column lies in the range of 0.2 to 0.4 that \citet{liu2025muonscalable} observe for AdamW, except Dion's at $\rho<1$.
}
\label{tab:update-rms}
\setlength{\tabcolsep}{4pt}
\begin{tabular}{@{}p{0.29\linewidth}lllr@{}}
\toprule
Raw update & $\|\mO\|_F^2$ & $s$ & RMS of $s\mO$ & GPT-2 \\
\midrule
Full-rank orthogonalized Muon & $\min(m,n)$ & $\tau\sqrt{\max(m,n)}$ & $\tau$ & 0.2 \\
Dion, $r$ orthonormal left / unit-norm right columns & $r$ & $\tau\sqrt{\max(m,n)}$ & $\tau\sqrt{r/\min(m,n)}$ & $0.2\sqrt{\rho}$ \\
RMNP, unit-norm rows & $m$ & $\tau\sqrt{\max(m,n)}$ & $\tau\sqrt{\max(m,n)/n}$ & 0.2 or 0.4$^{a}$ \\
\dobb, \dobbLtwo & $n$ & $\tau\sqrt{\max(m,n)}$ & $\tau\sqrt{\max(m,n)/m}$ & 0.2 or 0.4$^{b}$ \\
\dobbG, \dobbGLtwo & $m$ & $\tau\sqrt{\max(m,n)}$ & $\tau\sqrt{\max(m,n)/n}$ & 0.2 or 0.4$^{a}$ \\
\bottomrule
\end{tabular}\\[3pt]
{\footnotesize $^{a}$0.4 on the MLP input projection, for which $m=4n$; 0.2 on the other matrices.\\
$^{b}$0.4 on the MLP output projection, for which $n=4m$; 0.2 on the other matrices.}
\end{table}

The Muon row is the full-rank result of \citet{liu2025muonscalable}.
For an activation-side \dobb block, orthonormality gives
$\|\widehat{\mZ}_b\mU_b^\top\|_F^2=\|\widehat{\mZ}_b\|_F^2$.
Unit-norm coefficient columns give $|\mathcal{S}_b|$, and summing over blocks gives $n$.
Sign coefficients divided by $\sqrt{m}$ give the same count when every coefficient is nonzero.
Transposing the construction gives $m$ on the output side.
RMNP's $m$ unit-norm rows give squared norm $m$.
For Dion's raw factor product $\mL\mR^\top$, $\mL^\top\mL=\mI_r$ and unit-norm columns of $\mR$ imply
$\|\mL\mR^\top\|_F^2=\operatorname{tr}(\mR^\top\mR)=r$.
Mutual orthogonality of the right columns is not required.

With the shared update scale, Dion's update RMS therefore falls with rank as $\tau\sqrt{r/\min(m,n)}$.
Every GPT-2 matrix has $r=\rho\min(m,n)$ at the four rank fractions of Table~\ref{tab:quality}, so Dion's update RMS is $0.2\sqrt{\rho}$: 0.2, 0.1, 0.05, and 0.025 at $\rho=1$, $1/4$, $1/16$, and $1/64$.
For $\rho<1$, Dion's update RMS therefore lies below the range of 0.2 to 0.4.
We did not tune the learning rate separately for each Dion rank fraction or for each \dobb block count; all configurations at a model scale share the learning rate and weight decay.
In their comparison of rank fractions across model sizes, \citet{ahn2025dion} likewise use one fixed learning rate for Muon and for Dion at every rank fraction (their Appendix~E.1), and their update scaling factor depends on the matrix shape but not on the rank (their Section~3.1).
With the learning rate and weight decay unchanged, replacing Dion's update scale on GPT-2 Small with the rank-corrected update scale $\tau\sqrt{mn/r}$, which holds its update RMS at $\tau$ at every rank and equals the shared scale at full rank, changes the final validation loss from 3.038 to 3.034 at $\rho=1/4$, from 3.108 to 3.089 at $\rho=1/16$, and from 3.265 to 3.267 at $\rho=1/64$.
The correction recovers at most 18\% of the gap to the full-rank loss of 3.002 (0.019 of 0.106, at $\rho=1/16$), so at this scale Dion's loss increase at low rank arises mainly from the reduced rank and not from the smaller update RMS.

The update scale avoids a new global norm reduction during the update.
Finite-precision orthogonalization, normalization epsilons, and zero coefficients can change the attained RMS.

\subsection{Learning-rate scaling}
\label{app:lrfit}

\begin{figure}[ht]
\centering
\includegraphics[width=\linewidth]{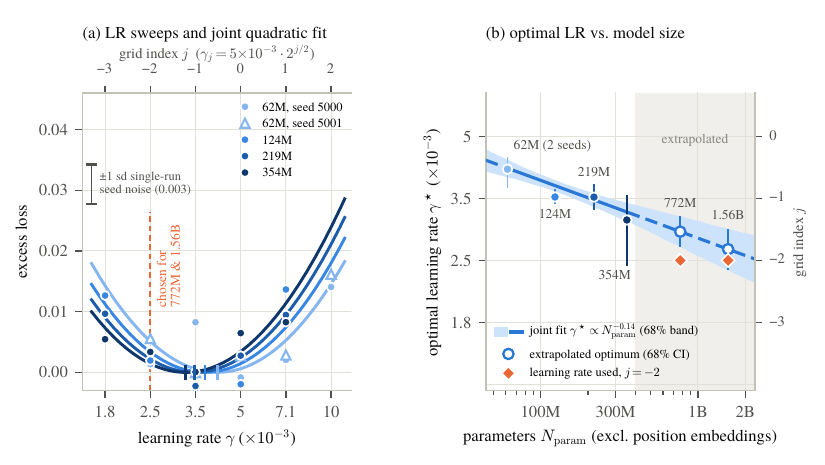}
\caption{Learning-rate scaling.
(a) Final validation loss relative to the fitted per-size minimum, with the shared-curvature parabolas of Equation~\ref{eq:lr-quad}; ticks at zero mark the fitted optima.
Both seeds at 62M are shown, and the bar marks the estimated run-to-run standard deviation.
(b) Optimal learning rate against parameter count.
Filled points are the per-size optima of Table~\ref{tab:lr-sweep} with 68\% intervals, and the line and band are the fitted law of Equation~\ref{eq:lr-fit} with its 68\% bootstrap band.
Each 68\% interval and band spans the 16th to 84th percentiles of its bootstrap distribution, which equals $\pm1$ standard deviation for a normal distribution.
The open circles are the extrapolated optima at 772M and 1.56B, and the diamonds mark the learning rate used for both, $\gamma=2.5\times10^{-3}$ ($j=-2$).
The shaded region lies beyond the largest swept size.}
\label{fig:lr-scaling}
\end{figure}

Let $\gamma$ denote the peak of the learning-rate schedule $\gamma_t$ and $\gamma^\star$ its optimal value.
We sweep $\gamma$ at four smaller model sizes, fit a power law for $\gamma^\star$, and extrapolate the fit to 772M and 1.56B.
The 124M, 354M, 772M, and 1.56B models have the architectures of GPT-2 Small, Medium, Large, and XL (Section~\ref{sec:setup}).
Table~\ref{tab:lr-sweep} lists the sizes and sweeps, and Figure~\ref{fig:lr-scaling} shows the fits.

\begin{table}[ht]
\centering
\small
\setlength{\tabcolsep}{2.4pt}
\caption{Model sizes and learning-rate sweeps.
$N_{\mathrm{param}}$ is the parameter count, including token embeddings and excluding position embeddings, and $D$ is the training token budget.
Sweeps use the grid of Equation~\ref{eq:lr-grid}, and grid ranges are inclusive.
For swept sizes, $\widehat{\gamma}^\star$ is the optimum of a quadratic in $j$ fitted to the five grid points centered on the sweep's best run, with both seeds at 62M; its 68\% interval comes from a parametric bootstrap at the measured seed noise.
For 772M and 1.56B, $\widehat{\gamma}^\star$ is the extrapolation of Equation~\ref{eq:lr-fit} with a 68\% residual-bootstrap interval; both models are trained at $\gamma=2.5\times10^{-3}$ ($j=-2$).
}
\label{tab:lr-sweep}
\begin{tabular}{rrrrrrcrc}
\toprule
$N_{\mathrm{param}}$ & Layers & Width & Heads & Steps & $D$ & Grid $j$ & Runs & $\widehat{\gamma}^\star$ ($10^{-3}$) \\
\midrule
\multicolumn{9}{l}{\emph{Swept sizes}} \\
62M & 6 & 640 & 10 & 5{,}000 & 2.46B & $-4\ldots4$ & $9+5^{a}$ & 4.16 [3.74, 4.45] \\
124M & 12 & 768 & 12 & 10{,}000 & 4.92B & $-4\ldots4$ & 9 & 3.56 [3.42, 3.72] \\
219M & 18 & 896 & 14 & 15{,}000 & 7.37B & $-3\ldots1$ & 5 & 3.56 [3.31, 3.83] \\
354M & 24 & 1024 & 16 & 20{,}000 & 9.83B & $-3\ldots1$ & 5 & 3.13 [2.43, 3.60] \\
\midrule
\multicolumn{9}{l}{\emph{Extrapolated sizes}} \\
772M & 36 & 1280 & 20 & 30{,}000 & 14.75B & n/a & n/a & 2.94 [2.70, 3.20] \\
1.56B & 48 & 1600 & 25 & 40{,}000 & 19.66B & n/a & n/a & 2.66 [2.37, 2.98] \\
\bottomrule
\end{tabular}\\[3pt]
{\footnotesize $^{a}$Second seed over $j=-2\ldots2$.}
\end{table}

\paragraph{Optimizer and update magnitude.}
The sweeps train all parameters with AdamW \citep{loshchilov2019adamw} at a fixed weight decay, so $\gamma$ is the only swept quantity.
In the standard parametrization, AdamW keeps the per-element update magnitude fixed as width grows.
The maximal update parametrization ($\mu$P) instead scales the Adam learning rate of hidden weights inversely with fan-in, which keeps the optimal learning rate stable across widths \citep{yang2022tensor}.
We therefore measure how $\gamma^\star$ depends on model size instead of assuming that it transfers.

\paragraph{Grid search.}
We sweep $\gamma$ over a geometric grid with ratio $\sqrt{2}$,
\begin{equation}
\gamma=5\times10^{-3}\cdot2^{j/2},\qquad j\in\mathbb{Z},
\label{eq:lr-grid}
\end{equation}
so adjacent grid points differ by a factor of $\sqrt{2}$ and points two steps apart by a factor of 2.
The 62M and 124M sweeps use $j=-4,\ldots,4$, from $1.25\times10^{-3}$ to $2\times10^{-2}$.
The 219M and 354M sweeps use $j=-3,\ldots,1$, from $1.77\times10^{-3}$ to $7.07\times10^{-3}$, because the two smaller sweeps had placed the optimum inside that window.
A second 62M sweep over $j=-2,\ldots,2$ with a different seed measures run-to-run noise.
Within a size, only $\gamma$ varies.

\paragraph{Training recipe.}
All sizes follow one recipe: head dimension 64, 491{,}520 tokens per optimizer step (480 sequences of 1024 tokens), and 0.41B training tokens per layer, so the token budget $D$ grows in proportion to depth.
The learning rate warms up linearly over the first 10\% of steps and then follows a cosine decay to 1\% of its peak; the weight decay $\lambda$ is 0.1.
Each run is scored by its final validation loss.
The 33 sweep runs use $2.0\times10^{20}$ FLOPs in total, counting $6N_{\mathrm{param}}D$ per run.

\paragraph{Noise floor.}
Across the five learning rates run with both seeds at 62M, the final losses differ by 0.005 RMS, giving an estimated run-to-run standard deviation of $\sigma\approx0.003$.
At 62M, the best grid point changes from $j=0$ to $j=-1$ between the two seeds.
We estimate $\gamma^\star$ by fitting the sweeps, with this noise estimate used for the per-size bootstrap intervals.

\paragraph{Joint fit.}
Let $\mathcal{L}(j;N_{\mathrm{param}})$ denote the final validation loss at grid index $j$ for a model with $N_{\mathrm{param}}$ parameters.
Near its minimum, $\mathcal{L}$ is well approximated by a quadratic in $\log\gamma$, that is, in $j$.
We fit all swept sizes jointly with a shared curvature $a$ and an optimum that follows a power law in $N_{\mathrm{param}}$:
\begin{align}
\mathcal{L}(j;N_{\mathrm{param}}) &= \mathcal{L}^\star(N_{\mathrm{param}})+a\,\big(j-j^\star(N_{\mathrm{param}})\big)^2, \label{eq:lr-quad}\\
j^\star(N_{\mathrm{param}}) &= j^\star_{\mathrm{ref}}-2\alpha\log_2\!\big(N_{\mathrm{param}}/N_{\mathrm{ref}}\big), \label{eq:lr-jstar}
\end{align}
where each size has a free offset $\mathcal{L}^\star$, and the reference size $N_{\mathrm{ref}}$ is 354M, the largest swept size, which anchors the fit inside the data.
The fit also determines $j^\star_{\mathrm{ref}}$, the optimal grid index at $N_{\mathrm{ref}}$, and the exponent $\alpha$.
Because $j=2\log_2(\gamma/5\times10^{-3})$, Equation~\ref{eq:lr-jstar} is equivalent to the power law
\begin{equation}
\gamma^\star(N_{\mathrm{param}})=\gamma^\star_{\mathrm{ref}}\,\big(N_{\mathrm{param}}/N_{\mathrm{ref}}\big)^{-\alpha},
\label{eq:lr-powerlaw}
\end{equation}
where $\gamma^\star_{\mathrm{ref}}=5\times10^{-3}\cdot2^{j^\star_{\mathrm{ref}}/2}$ is the optimal learning rate at $N_{\mathrm{ref}}$.
The parabola is a local approximation, and the sweeps depart from it beyond about two grid steps from the optimum.
Each size therefore contributes only the five grid points centered on its best run, with both seeds at 62M, which gives 25 points for 7 parameters.
We fit by nonlinear least squares and derive confidence intervals from 600 residual-bootstrap resamples of the joint fit.

\paragraph{Fitted law.}
The joint fit gives
\begin{equation}
\gamma^\star(N_{\mathrm{param}})=3.28\times10^{-3}\left(\frac{N_{\mathrm{param}}}{354\,\mathrm{M}}\right)^{-0.140},
\label{eq:lr-fit}
\end{equation}
with $\alpha=0.140$ (68\% interval $[0.096, 0.186]$; 95\% interval $[0.051, 0.224]$) and a 68\% interval of $[3.11, 3.45]\times10^{-3}$ for $\gamma^\star_{\mathrm{ref}}$.
The shared curvature is $a=0.0023$ per squared grid step, so a learning rate off by a factor of $\sqrt{2}$ raises the final loss by about 0.002, and one off by a factor of 2 raises it by about 0.009.
The residual standard deviation of the fit is 0.003, which matches the measured seed noise, so the model describes the sweeps to within noise.
Pooling the sizes also keeps a near-tie within one sweep from deciding the fit; at 124M, for example, $j=-1$ and $j=0$ differ by less than 0.001.

\paragraph{Extrapolation to 772M and 1.56B.}
Rounding the extrapolated optima in Table~\ref{tab:lr-sweep} to the nearest grid point gives $j=-2$ for both sizes, so both models use $\gamma=2.5\times10^{-3}$.
Under the fitted quadratic, this rounding increases the predicted loss by less than 0.001.

\paragraph{Limitations.}
The 772M and 1.56B predictions extrapolate beyond the swept range.
Because width, depth, and data grow together, Equation~\ref{eq:lr-fit} describes this recipe and cannot attribute the change in $\gamma^\star$ to any single factor.
In particular, it combines the effect of model size with that of the training horizon, which also shifts the optimal learning rate \citep{bjorck2025scaling}.
Batch size, warmup fraction, and schedule are held fixed, and the fit applies only to these settings.

\section{Ablations}
\label{app:ablations}

\paragraph{Configurations for the basis and normalization ablations.}
We use three alternative configurations to isolate the choices of activation-side basis and sign normalization in \dobb.
\dobbLtwo retains the activation-side basis and normalizes every coefficient column to unit Euclidean norm, with a zero column mapped to zero.
For an output-side basis, we partition the output coordinates and construct $\mC_{t,b}$ from $\mD_{t,\mathcal{S}_b,:}$.
We project the transposed momentum slice $\mM_{t,\mathcal{S}_b,:}^{\top}$, normalize its coefficients, reconstruct, and transpose back.
\dobbG uses this basis with sign normalization and denominator $\sqrt{n}$; \dobbGLtwo instead normalizes each coefficient column to unit Euclidean norm.
The four configurations are summarized below.
\begin{center}
\begin{tabular}{lll}
\toprule
Basis source & Sign normalization & Column normalization \\
\midrule
Input second moment $\mA_t$ & \dobb & \dobbLtwo \\
Output-gradient second moment $\mG_t$ & \dobbG & \dobbGLtwo \\
\bottomrule
\end{tabular}
\end{center}
All four configurations use the update scale $s=\tau\sqrt{\max(m,n)}$ of Equation~\ref{eq:dobb-scale}, and Appendix~\ref{app:update-rms} gives the resulting update RMS of each.

Unless stated otherwise, each ablation changes only its named component and retains the principal algorithm settings, update scale, weight decay, token budget, and evaluation procedure.
Each ablation also uses the principal comparison's learning rate at its model size, selected as described in Appendix~\ref{app:protocol}.
Unless a subsection states otherwise, each ablation trains GPT-2 Small with $B=1$ and $T=9$ for 10{,}000 iterations with seed 5000, using data parallelism on four H100 GPUs, which replicates rather than shards the model.
Final loss denotes final validation loss.
Unless a caption states otherwise, iteration times include second-moment capture and are reported in milliseconds.
Non-refresh and Refresh denote the median full-iteration times $\theta_{\mathrm{N}}$ and $\theta_{\mathrm{R}}$ on iterations without and with a basis refresh; Amortized denotes $(\theta_{\mathrm{R}}+(T-1)\theta_{\mathrm{N}})/T$.

\subsection{Sign versus \texorpdfstring{$\ell_2$}{l2} normalization}
\label{app:norm}

\dobb and \dobbLtwo share the activation-side basis and differ only in the coefficient normalization $\mathcal{N}$ of Equation~\ref{eq:dobb-update}: \dobb applies the sign, and \dobbLtwo scales each coefficient column to unit $\ell_2$ norm.
This ablation departs from the default setting above: it uses the principal comparison's runs of both variants on GPT-2 Small, Medium, Large, and XL at $B\in\{1,4,16,64\}$, with the training budgets of Table~\ref{tab:quality}.

Table~\ref{tab:norm} shows that sign normalization gives the lower final loss in all sixteen matched comparisons, by 0.008 to 0.026.
On Small, the margin decreases from 0.026 at $B=1$ to 0.008 at $B=64$: \dobb's loss rises by 0.020 over this range, while \dobbLtwo's loss stays between 3.036 and 3.045.
On Medium, Large, and XL, the margin stays between 0.014 and 0.019, between 0.019 and 0.021, and between 0.012 and 0.016, respectively.
At these three scales, the two variants change by similar amounts from $B=1$ to $B=64$: by 0.007 and 0.006 on Medium, by 0.021 and 0.020 on Large, and by 0.021 and 0.022 on XL.
Against Dion at rank fraction $1/B$ in Table~\ref{tab:quality}, \dobbLtwo has lower loss at $B=16$ and $B=64$ on all four scales.
At $B=4$, \dobbLtwo is 0.001 below Dion on Small and 0.006, 0.013, and 0.002 above it on Medium, Large, and XL.

The two normalizations also differ in derived cross-shard communication.
Sign normalization acts on each coefficient independently, so when each block lies within one shard, \dobb adds no optimizer-specific cross-shard payload beyond the data-parallel synchronization of its block second moments.
Column normalization sums squares over each coefficient column, so \dobbLtwo reduces $n$ scalars per matrix when the output axis is sharded (Table~\ref{tab:comm-ablations}).

\begin{table}[ht]
\centering
\small
\caption{Normalization ablation for the activation-side basis: final validation loss for \dobb (sign) and \dobbLtwo (column $\ell_2$).
The \dobb columns repeat Table~\ref{tab:quality}; $\Delta$ is the \dobbLtwo loss minus the \dobb loss.
}
\label{tab:norm}
\setlength{\tabcolsep}{3.5pt}
\begin{tabular}{rrrrrrrrrrrrr}
\toprule
 & \multicolumn{3}{c}{Small} & \multicolumn{3}{c}{Medium} & \multicolumn{3}{c}{Large} & \multicolumn{3}{c}{XL} \\
\cmidrule(lr){2-4}\cmidrule(lr){5-7}\cmidrule(lr){8-10}\cmidrule(lr){11-13}
$B$ & \dobb & \dobbLtwo & $\Delta$ & \dobb & \dobbLtwo & $\Delta$ & \dobb & \dobbLtwo & $\Delta$ & \dobb & \dobbLtwo & $\Delta$ \\
\midrule
1 & 3.017 & 3.043 & 0.026 & 2.735 & 2.753 & 0.018 & 2.579 & 2.599 & 0.020 & 2.461 & 2.476 & 0.015 \\
4 & 3.024 & 3.037 & 0.013 & 2.741 & 2.760 & 0.019 & 2.588 & 2.609 & 0.021 & 2.471 & 2.483 & 0.012 \\
16 & 3.026 & 3.036 & 0.010 & 2.746 & 2.760 & 0.014 & 2.595 & 2.615 & 0.020 & 2.476 & 2.491 & 0.015 \\
64 & 3.037 & 3.045 & 0.008 & 2.742 & 2.759 & 0.017 & 2.600 & 2.619 & 0.019 & 2.482 & 2.498 & 0.016 \\
\bottomrule
\end{tabular}
\end{table}

\subsection{Input-activation versus output-gradient bases}
\label{app:basis}

The four Dobb variants separate the basis source from the coefficient normalization.
Appendix~\ref{app:norm} compares the activation-side variants across scales; this ablation adds \dobbG and \dobbGLtwo at $B\in\{1,4,16,64\}$.
We compare A versus G at each fixed normalization and block count, then compare the normalizations at each fixed basis source.

Table~\ref{tab:basis} shows that the activation basis gives the lower final loss in all eight matched comparisons.
\dobbG exceeds \dobb by 0.035 to 0.054, and \dobbGLtwo exceeds \dobbLtwo by 0.034 to 0.039.
The two groups do not overlap: the highest activation-side loss, 3.045, is 0.022 below the lowest output-gradient loss, 3.067.
Every normalization difference, at most 0.026, is smaller than every basis-source difference.
With the activation basis, sign normalization gives the lower loss at every $B$, as Appendix~\ref{app:norm} reports for all four scales.
With the output-gradient basis, sign normalization gives lower loss by 0.006 at $B=1$ and 0.011 at $B=64$, while at $B=4$ and $B=16$ the two normalizations differ by at most 0.004.

\begin{table}[ht]
\centering
\small
\caption{Basis-source and normalization ablation.
The activation-side columns repeat the Small entries of Table~\ref{tab:norm}.
}
\label{tab:basis}
\begin{tabular}{rrrrr}
\toprule
 & \multicolumn{2}{c}{Activation basis} & \multicolumn{2}{c}{Output-gradient basis} \\
\cmidrule(lr){2-3}\cmidrule(lr){4-5}
$B$ & \dobb & \dobbLtwo & \dobbG & \dobbGLtwo \\
\midrule
1 & 3.017 & 3.043 & 3.071 & 3.077 \\
4 & 3.024 & 3.037 & 3.067 & 3.071 \\
16 & 3.026 & 3.036 & 3.076 & 3.075 \\
64 & 3.037 & 3.045 & 3.072 & 3.083 \\
\bottomrule
\end{tabular}
\end{table}

\subsection{Gradient-derived second moments}
\label{app:gradcov}

We also consider a hook-free basis constructed from the mini-batch weight gradient itself.
The input- and output-side gradient Gram matrices are
\begin{equation}
\begin{aligned}
\mS_t^{\mathrm{in}}
 &= \big[\nabla_{\mW}f_{\mathcal{B}_t}(\mW_t)\big]^\top
    \big[\nabla_{\mW}f_{\mathcal{B}_t}(\mW_t)\big],\\
\mS_t^{\mathrm{out}}
 &= \big[\nabla_{\mW}f_{\mathcal{B}_t}(\mW_t)\big]
    \big[\nabla_{\mW}f_{\mathcal{B}_t}(\mW_t)\big]^\top.
\end{aligned}
\end{equation}
These gradient Gram matrices are the type of statistics accumulated by Shampoo and by SOAP, which runs Adam in the Shampoo eigenbasis \citep{gupta2018shampoo,vyas2024soap}.
A one-sided block eigenbasis uses sign or column normalization.
The input-side construction stores a block-diagonal second moment and its basis, $2n^2/B$ scalars, and each refresh factorizes $B$ blocks of width $n/B$.
With $\ell_\infty$ (sign) normalization, the output-side construction matches the one-sided simplification of SPlus in the unified view of rotated steepest descent \citep{gong2026aro}, and the input-side construction is its counterpart on the input side; SPlus itself rotates by both Shampoo eigenbases and adds shape-aware scaling and iterate averaging \citep{frans2025splus}.
The ablation compares final loss for each source under both normalizations.
The two kinds of source also differ in sampling: each gradient Gram matrix enters its exponential average at every iteration and uses the full mini-batch gradient, whereas the capture hooks record only the first micro-batch on refresh iterations.
The comparison changes both the statistic and its sampling schedule.

Table~\ref{tab:gradcov} shows that every input-side source has lower final loss than every output-side source under both normalizations.
Under $\ell_\infty$ normalization, final loss rises from 3.017 for the activation second moment to 3.027 and 3.055 for the input- and output-side gradient Gram matrices, and to 3.071 for the output-gradient second moment.
Under $\ell_2$ normalization the ordering is the same, except that the activation second moment and the input-side gradient Gram matrix differ by 0.001 (3.043 and 3.042).
On the input side, the activation second moment has 0.010 lower final loss than the gradient Gram matrix under sign normalization.
On the output side, the gradient Gram matrix has lower final loss than the output-gradient second moment by 0.016 under $\ell_\infty$ and by 0.013 under $\ell_2$ normalization; this difference combines the statistic with its denser sampling.

\begin{table}[ht]
\centering
\small
\caption{Second-moment-source ablation.
}
\label{tab:gradcov}
\setlength{\tabcolsep}{5pt}
\begin{tabular}{llr}
\toprule
Norm & Source & Final loss \\
\midrule
$\ell_\infty$ & Activation second moment (\dobb) & 3.017 \\
 & Gradient Gram, input side & 3.027 \\
 & Gradient Gram, output side & 3.055 \\
 & Output-gradient second moment (\dobbG) & 3.071 \\
\midrule
$\ell_2$ & Activation second moment (\dobbLtwo) & 3.043 \\
 & Gradient Gram, input side & 3.042 \\
 & Gradient Gram, output side & 3.064 \\
 & Output-gradient second moment (\dobbGLtwo) & 3.077 \\
\bottomrule
\end{tabular}
\end{table}

\subsection{Basis-refresh period}
\label{app:refresh}

The principal period is $T=9$.
We compare final loss and iteration time at $T=1,4,9,16,100$.
Because capture occurs only on refresh iterations, $T$ determines how long the basis is reused and, over a fixed training budget, how many micro-batches contribute to the second-moment accumulator.

Table~\ref{tab:refresh} shows that refreshing at every iteration gives the lowest final loss, 3.005, which is 0.012 below the 3.017 of the principal $T=9$.
From $T=4$ to $T=100$, final loss stays between 3.014 and 3.017 without a monotone trend.
Relative to $T=1$, the principal $T=9$ reduces the amortized iteration time from 762.5 to 403.6~ms, or by 47\%.
For fixed refresh and non-refresh times, the amortized refresh overhead decreases as $1/T$.
The measured amortized iteration time falls from 403.6~ms at $T=9$ to 384.8~ms at $T=16$ and to 363.4~ms at $T=100$, so the longer periods lower it by at most 40.2~ms and change final loss by at most 0.002.

\begin{table}[ht]
\centering
\small
\caption{Refresh-period ablation for \dobb.
Full iteration reports the amortized time.
}
\label{tab:refresh}
\begin{tabular}{rrr}
\toprule
$T$ & Final loss & Full iteration (ms) \\
\midrule
1 & 3.005 & 762.5 \\
4 & 3.014 & 459.3 \\
9 & 3.017 & 403.6 \\
16 & 3.017 & 384.8 \\
100 & 3.015 & 363.4 \\
\bottomrule
\end{tabular}
\end{table}

\subsection{Orthogonal iterations per refresh}
\label{app:orthiter}

Let $k$ denote the number of warm-started orthogonal iterations per refresh; Algorithm~\ref{alg:eig} uses $k=1$.
We compare this principal setting with $k=2$ and $k=3$ and with an exact eigendecomposition at every refresh.
Each additional iteration repeats the product $(\mC_{t,b}+\varepsilon\mI)\mU_b$ and its QR orthonormalization, with $\varepsilon$ selected as in Algorithm~\ref{alg:eig}, so the derived refresh arithmetic grows linearly in $k$.
The exact variant computes the eigendecomposition of $\mC_{t,b}+\varepsilon\mI$ in fp32 at every refresh and discards the warm start.
The block update is invariant to the sign and order of the basis columns, so the sign and ordering conventions of the eigensolver do not affect it.

Table~\ref{tab:orthiter} shows that final loss varies by 0.004 across the four settings, from 3.015 for the exact eigendecomposition to 3.019 at $k=2$, and does not decrease monotonically with $k$.
The extra cost falls on refresh iterations.
Non-refresh iterations take 358.5 to 360.1~ms in all four settings, while each additional orthogonal iteration adds 375 to 381~ms to a refresh iteration and the exact eigendecomposition adds 632~ms.
At $T=9$ these refresh costs raise the amortized iteration time by 10.6\%, 20.6\%, and 17.5\% for $k=2$, $k=3$, and the exact variant.
The principal $k=1$ setting has the lowest amortized iteration time, 403.6~ms; exact eigendecomposition takes 474.4~ms and lowers loss from 3.017 to 3.015.

\begin{table}[ht]
\centering
\small
\caption{Orthogonal-iteration ablation for \dobb.
}
\label{tab:orthiter}
\begin{tabular}{lrrrr}
\toprule
 & & \multicolumn{3}{c}{Full iteration (ms)} \\
\cmidrule(lr){3-5}
Basis refresh & Final loss & Non-refresh & Refresh & Amortized \\
\midrule
$k=1$ (principal) & 3.017 & 359.5 & 756.5 & 403.6 \\
$k=2$ & 3.019 & 359.9 & 1137.8 & 446.4 \\
$k=3$ & 3.017 & 358.5 & 1512.3 & 486.7 \\
Exact eigendecomposition & 3.015 & 360.1 & 1388.9 & 474.4 \\
\bottomrule
\end{tabular}
\end{table}

\subsection{Householder versus Cholesky QR}
\label{app:cholqr}

This ablation compares Householder QR with the Cholesky QR of Algorithm~\ref{alg:cholqr} at $\omega=10^{-2}$ in \dobb at $B=1$ and $B=4$.

Cholesky QR lowers the iteration time with little change in final loss.
Table~\ref{tab:cholqr} shows that the QR routine changes final loss by at most 0.002: Cholesky QR gives 3.016 against 3.017 for Householder QR at $B=1$, and 3.022 against 3.024 at $B=4$.
At iteration 1000, Cholesky QR has 0.012 higher validation loss at both block counts; from iteration 2000 onward, the two routines differ by at most 0.005.
The time saving falls primarily on refresh iterations: their median time decreases from 760.3 to 486.3~ms at $B=1$ and from 604.7 to 431.8~ms at $B=4$, while the non-refresh medians change by at most 3.5~ms.
At $T=9$, Cholesky QR lowers the amortized iteration time from 409.1 to 375.6~ms, a reduction of 33.5~ms (8.2\%), at $B=1$ and from 417.6 to 397.5~ms, a reduction of 20.1~ms (4.8\%), at $B=4$.

\begin{table}[ht]
\centering
\small
\caption{QR-routine ablation for \dobb.
}
\label{tab:cholqr}
\begin{tabular}{rlrrrr}
\toprule
 & & & \multicolumn{3}{c}{Full iteration (ms)} \\
\cmidrule(lr){4-6}
$B$ & QR routine & Final loss & Non-refresh & Refresh & Amortized \\
\midrule
1 & Householder & 3.017 & 365.2 & 760.3 & 409.1 \\
 & Cholesky & 3.016 & 361.7 & 486.3 & 375.6 \\
\midrule
4 & Householder & 3.024 & 394.2 & 604.7 & 417.6 \\
 & Cholesky & 3.022 & 393.2 & 431.8 & 397.5 \\
\bottomrule
\end{tabular}
\end{table}

\subsection{First-micro-batch capture}
\label{app:micro0}

The principal policy, first-micro-batch capture, forms the second moment on each refresh iteration from the first micro-batch alone.
The alternative, all-micro-batch capture, forms a token-weighted second moment from every micro-batch of that iteration.
Both policies use the full mini-batch stochastic gradient for momentum, so they differ only in the data from which the second moment is estimated.
Each iteration accumulates gradients over eight micro-batches of 15 sequences on each of the four GPUs, so first-micro-batch capture pools 60 of the iteration's 480 sequences across GPUs.
We run the comparison for \dobb and \dobbG.

Table~\ref{tab:micro0} shows that capturing every micro-batch does not lower final loss.
For \dobb it gives 3.021 against 3.017 with first-micro-batch capture, and for \dobbG both policies give 3.071.
Capturing every micro-batch raises the median refresh iteration time by 132.4~ms for \dobb and by 59.6~ms for \dobbG, while the non-refresh medians differ by at most 0.9~ms.
At $T=9$ this raises the amortized iteration time by 15.5~ms (3.8\%) for \dobb and by 6.5~ms (1.5\%) for \dobbG.

\begin{table}[ht]
\centering
\small
\caption{Capture-policy ablation for \dobb and \dobbG.
}
\label{tab:micro0}
\begin{tabular}{llrrrr}
\toprule
 & & & \multicolumn{3}{c}{Full iteration (ms)} \\
\cmidrule(lr){4-6}
Method & Capture policy & Final loss & Non-refresh & Refresh & Amortized \\
\midrule
\dobb & First micro-batch & 3.017 & 360.9 & 755.7 & 404.8 \\
 & All micro-batches & 3.021 & 361.8 & 888.1 & 420.3 \\
\midrule
\dobbG & First micro-batch & 3.071 & 370.4 & 835.1 & 422.0 \\
 & All micro-batches & 3.071 & 370.2 & 894.7 & 428.5 \\
\bottomrule
\end{tabular}
\end{table}

\subsection{Second-moment and basis precision}
\label{app:precision}

The principal setting accumulates the block second moments in fp32 and stores the bases in bf16.
The implementation forms each captured outer product in the activation dtype and casts it to the accumulator dtype; at each refresh it casts the accumulator to fp32 before adding the shift and factorizing.
Accumulation precision therefore affects the exponential average and its storage, not the factorization arithmetic, while basis precision affects the projection and reconstruction in Equation~\ref{eq:dobb-update}.
For side $A$ the accumulator and the basis each hold $n^2/B$ entries per matrix, so each bf16 choice halves that buffer.
We compare the four combinations of accumulation and basis precision.

Table~\ref{tab:precision} shows that fp32 accumulation gives lower final loss than bf16 accumulation at both basis precisions: 3.023 against 3.028 with fp32 bases, and 3.017 against 3.027 with bf16 bases.
Basis precision changes final loss by 0.006 at fp32 accumulation and by 0.001 at bf16 accumulation, in both cases in favor of bf16 bases.
The principal combination, fp32 accumulation with bf16 bases, has the lowest final loss of the four.
Relative to the principal setting, bf16 accumulation reduces the derived second-moment and basis storage from $6n^2/B$ to $4n^2/B$ bytes per matrix and raises final loss by 0.010.

\begin{table}[ht]
\centering
\small
\caption{Precision ablation for \dobb.
Median iteration is the median full-iteration time in milliseconds on iterations without an eigenbasis refresh.
The storage column gives the second-moment and basis storage per matrix and is derived from buffer sizes, not measured.
}
\label{tab:precision}
\begin{tabular}{llrrr}
\toprule
Accumulation & Basis & Final loss & Median iteration (ms) & Storage per matrix (bytes) \\
\midrule
fp32 & fp32 & 3.023 & 361.2 & $8n^2/B$ \\
fp32 & bf16 (principal) & 3.017 & 360.4 & $6n^2/B$ \\
bf16 & fp32 & 3.028 & 361.0 & $6n^2/B$ \\
bf16 & bf16 & 3.027 & 360.9 & $4n^2/B$ \\
\bottomrule
\end{tabular}
\end{table}

\subsection{Nesterov momentum}
\label{app:nesterov}
The principal \dobb definition uses ordinary momentum with Nesterov disabled.
The ablation compares the update input $\mM_t$ with the Nesterov input
$\nabla_{\mW}f_{\mathcal{B}_t}(\mW_t)+\mu\mM_t$ after the same momentum accumulation.
It tests whether the additional gradient contribution changes final quality under the shared update scale.
The principal comparison disables Nesterov momentum for \dobb, Muon, and RMNP at every model size.
We test \dobb at $B=1$, 4, and 16, and the Muon and RMNP baselines.

Table~\ref{tab:nesterov} shows that the effect of Nesterov momentum on \dobb changes sign with $B$.
It lowers final loss by 0.003 at $B=1$ (3.017 to 3.014) and by 0.001 at $B=4$, and raises it by 0.004 at $B=16$ (3.026 to 3.030).
For the baselines, Nesterov momentum leaves the final loss of Muon at 3.004 and lowers that of RMNP by 0.011, from 3.083 to 3.072.
Nesterov momentum changes the amortized iteration time of \dobb by at most 0.7\%.

\begin{table}[ht]
\centering
\small
\caption{Nesterov ablation.
}
\label{tab:nesterov}
\begin{tabular}{lrrr}
\toprule
Method & $B$ & Off & On \\
\midrule
\dobb & 1 & 3.017 & 3.014 \\
\dobb & 4 & 3.024 & 3.023 \\
\dobb & 16 & 3.026 & 3.030 \\
\midrule
Muon & n/a & 3.004 & 3.004 \\
RMNP & n/a & 3.083 & 3.072 \\
\bottomrule
\end{tabular}
\end{table}

\section{Details of the distributed iteration-time experiments}
\label{app:distributed}

This appendix gives the settings and the full timing breakdown of the four campaigns summarized in Table~\ref{tab:fsdp}.
Three campaigns used our implementation, built on PyTorch FSDP2 with DTensor-based TP \citep{liang2025torchtitan}: GPT-2 Large and XL on A5000 GPUs (Appendix~\ref{app:distributed-a5000}), GPT-2 XL on H100 GPUs, and GPT-2 3B, 6B, and 8B on H100 GPUs (Appendix~\ref{app:distributed-h100}).
The fourth campaign timed Qwen3-8B and Qwen3-32B on H100 GPUs with an implementation built on the open-source Dion codebase \citep{ahn2025dion}, which we call the Dion-based implementation (Appendix~\ref{app:distributed-qwen}).
It also uses PyTorch FSDP2 with DTensor-based TP.
For Qwen3, this appendix also reports timings for the output-gradient basis with column normalization, \dobbGLtwo, defined in Algorithm~\ref{alg:dobb-full}.
All four campaigns used NCCL over TCP sockets between nodes.
The optimizer-call times are measured, and the pooled iteration times are derived from measured means; the communication counts in Section~\ref{sec:comm} are analytical, and the tables report times only.
Pooled iteration time is the comparison metric (Appendix~\ref{app:pooled}).

In every campaign, \dobb refreshes its bases every $T=9$ iterations.
Both Qwen3 series use the fixed-count partition of Appendix~\ref{app:algorithm}, with $B=\FS$ blocks per weight matrix.
Each Qwen3 series pairs Dion's rank fraction with a width of $1024/\FS$, following the width--rank match of Section~\ref{sec:comm}.
In the GPT-2 campaigns, \dobb's blocks follow the shards, so each block lies within one shard.

Our implementation applies the matrix optimizers to the two-dimensional weights of the transformer blocks and AdamW to the remaining parameters.
It runs a distributed Muon \citep{jordan2024muon} that orthogonalizes each whole weight matrix with Newton--Schulz iterations.
Every GPT-2 run started from a random initialization, used a global batch of 480 OpenWebText sequences of 1024 tokens, and kept the gathered FSDP weights across accumulation micro-steps.

Our implementation compiles the model with \texttt{torch.compile}, whereas the Dion-based implementation disables model compilation; both log timings on one rank.

\subsection{Pooled iteration time}
\label{app:pooled}

At a given model and device mesh, the forward-backward pass performs the same model computation for every optimizer; for \dobb it also contains the second-moment capture on refresh iterations and the reduction of the captured second moments, which our implementation launches from the capture hook behind the backward pass.
We measured the run-to-run variation of forward-backward time on GPT-2 Large at $\FS=10$, $\TP=2$ on 20 A5000 GPUs in three allocations, each running AdamW, Muon, Dion $r=400$, and \dobb once in each of four rounds, for 48 runs of 30 timed iterations; the run order was fixed in one allocation and balanced over position and predecessor by a Williams Latin square in the other two.
Table~\ref{tab:fwdbwd-noise} compares the forward-backward pass means and their run-to-run variation.
\dobb's mean forward-backward time of 7740~ms lies between the lowest and highest baseline means, 7713~ms for Dion and 7788~ms for Muon.

\begin{table}[t]
\centering
\caption{Forward-backward time on GPT-2 Large at $\FS=10$, $\TP=2$ on 20 A5000 GPUs.
Each optimizer has 12 runs across three allocations, with 30 timed iterations per run.
The standard deviation and range width describe the 12 run means.
All times are in milliseconds.}
\label{tab:fwdbwd-noise}
\begin{tabular}{lrrr}
\toprule
Optimizer & Mean & Run-to-run SD & Range width \\
\midrule
AdamW & 7723 & 22.8 & 77 \\
Muon & 7788 & 26.9 & 81 \\
Dion $r=400$ & 7713 & 16.0 & 56 \\
\dobb & 7740 & 15.3 & 48 \\
\bottomrule
\end{tabular}
\end{table}

After round and method effects within each allocation are removed, a run's mean forward-backward time has a standard deviation of 16.6~ms (0.21\%).
In the 12 pairs of \dobb and AdamW runs from the same round, \dobb's forward-backward mean exceeds AdamW's by 16.8~ms (0.22\%) on average, with a 95\% confidence interval of 2.4 to 31.3~ms; this overhead is comparable to the run-to-run standard deviation and small against the variation of a single pair's difference, about $\pm51$~ms at the 95\% level.
\dobb's mean iteration time exceeds AdamW's by 13.6~ms (0.18\%), with a 95\% confidence interval of $-1.1$ to 28.4~ms that includes zero.
Muon's and Dion's mean iteration times exceed AdamW's by 672~ms (8.68\%) and 1209~ms (15.61\%).
Thus, \dobb's total iteration-time overhead is indistinguishable from network timing noise in these runs.
To compare optimizer-step costs more fairly in the presence of this noise, we compute pooled iteration times by adding each method's measured optimizer-call mean to shared estimates of forward-backward time and the remaining iteration time, each averaged over all reported runs of the same configuration.
Each method then adds its own measured optimizer-call mean, so within a configuration, pooled iteration times differ only by the optimizer call, and pooling excludes optimizer-specific work inside forward-backward, including \dobb's capture and reduction.
In the Qwen3-32B runs, the Dion-based TP implementation reduces \dobb's second moments inside the optimizer call instead (Appendix~\ref{app:distributed-qwen}), so there pooling excludes only the capture.
For \dobb, the optimizer-call mean is averaged over one refresh cycle, $(\text{Refresh}+8\cdot\text{Non-refresh})/9$.

\subsection{Work that can leave the optimizer call}
\label{app:off-step}

An operation can leave the optimizer call once its inputs are available if it does not require the current mini-batch weight gradient.
Table~\ref{tab:off-step} classifies the operations of the compared optimizers and of Shampoo, SOAP, and K-FAC by this criterion.
Every operation of AdamW, Muon, and the standard Dion update reads the gradient.
The lazy variant of Dion refreshes its right factor only every $k$ steps, and the CPU variant computes this refresh on CPUs from the momentum of earlier steps while forward-backward runs on the GPUs \citep[Section~5.1]{ahn2025dion}.
Shampoo and SOAP accumulate their statistics from the gradient, so their inverse roots or eigenbases can leave the optimizer call only if they use statistics from earlier steps.
SOAP recomputes its eigenbases at the end of a step and first uses them in the next \citep[Algorithm~3]{vyas2024soap}, and \citet{shi2023distshampoo} describe a JAX implementation of Shampoo that overlaps the root-inverse computation, offloaded to CPUs, with updates that use stale roots.
K-FAC and \dobb estimate their statistics from activations and output gradients instead, so neither their statistics nor the inverses or bases derived from them read the gradient of any step.
\citet{martens2015kfac} obtain the K-FAC output-gradient factor from an additional backward pass with sampled targets (their Section~5), recompute the inverses every 20 iterations in their experiments, and note that these computations can run simultaneously with the gradient computation (their Section~8).

\begin{table}[t]
\centering
\small
\caption{Dependence of optimizer operations on the current gradient, the mini-batch gradient of a weight matrix after its reduction across data-parallel ranks.
Step: the operation reads the gradient and runs in the optimizer call.
Fwd/bwd: it reads activations or output gradients and runs during forward-backward.
Movable: it reads no gradient and can run before the optimizer call of the same iteration, once its inputs exist.
Lagged: it reads statistics or momentum that include gradients, so it can leave the optimizer call only by using values from earlier steps.
The first Dion row describes the standard algorithm used in our experiments.
Decoupled weight decay reads only the weights and is omitted.
$^{a}$Our GPT-2 implementation launches the reduction behind the backward pass and runs the refresh inside the optimizer call.}
\label{tab:off-step}
\setlength{\tabcolsep}{4pt}
\begin{tabular}{@{}l>{\raggedright\arraybackslash}p{0.33\linewidth}>{\raggedright\arraybackslash}p{0.20\linewidth}ll@{}}
\toprule
Optimizer & Operation & Reads & Frequency & Placement \\
\midrule
AdamW & First and second moments; element-wise update & Gradient & Every step & Step \\
\midrule
Muon & Momentum; Newton--Schulz orthogonalization & Gradient & Every step & Step \\
\midrule
Dion & Momentum; power iteration with orthonormalization; column normalization; error feedback & Gradient & Every step & Step \\
 & Right-factor refresh in the lazy and CPU variants & Momentum & Periodic & Lagged \\
\midrule
Shampoo, SOAP & Gradient Gram matrices & Gradient & Every step & Step \\
 & Inverse roots (Shampoo) or eigenbases (SOAP) & Gram matrices & Every $f$ steps & Lagged \\
 & Preconditioned update, or Adam in the eigenbasis & Gradient & Every step & Step \\
\midrule
K-FAC & Activation and output-gradient factors & Activations, output gradients & Every step & Fwd/bwd \\
 & Damped factor inverses & Factors & Every $T_3$ steps & Movable \\
 & Preconditioned gradient & Gradient & Every step & Step \\
\midrule
\dobb & Capture and block second moments & Activations & Every $T$ steps & Fwd/bwd \\
 & Second-moment reduction & Block second moments & Every $T$ steps & Movable$^{a}$ \\
 & Basis refresh & Block second moments, previous basis & Every $T$ steps & Movable$^{a}$ \\
 & Momentum; projection, sign, and reconstruction & Gradient & Every step & Step \\
\bottomrule
\end{tabular}
\end{table}

\paragraph{Scheduling \dobb's work outside the optimizer call.}
The refresh reads only the block second moments and the previous basis (Algorithm~\ref{alg:eig}), yet both implementations run it inside the optimizer call on refresh iterations.
For the activation-side basis, the refresh of a weight matrix could start once its second moment has been captured and reduced, while forward-backward computation continues for other layers.
Work moved out of the optimizer call still occupies the GPUs, so it shortens the iteration only to the extent that it fills intervals in which they would otherwise wait; we did not measure this overlap.
\citet{osawa2023pipefisher} assign the curvature and inversion work of K-FAC to the bubbles of pipeline-parallel schedules and refresh the curvature once every few steps, so that preconditioning is the only added computation.
Explicit scheduling of \dobb's capture, reduction, and basis refresh into idle intervals remains future work.

\subsection{GPT-2 on A5000 GPUs}
\label{app:distributed-a5000}

The A5000 campaign ran on nodes with two RTX A5000 GPUs each.
GPT-2 Large ran at $\DP\times\FS\times\TP=1\times10\times2$ on 10 nodes without activation checkpointing, with micro-batches of 8 sequences and 6 accumulation micro-steps.
GPT-2 XL ran on 16 nodes at $1\times32\times1$, $2\times16\times1$, and $2\times8\times2$, with activation checkpointing of every transformer block and the largest micro-batches that fit: 15 sequences, with one accumulation micro-step at $\TP=1$ and two at $\TP=2$.
The $2\times8\times2$ mesh uses 20 attention heads instead of 25, because 25 heads do not split over two TP ranks; the matrix shapes and the parameter count are unchanged.
Each run lasted 40 iterations, of which the first 10 were untimed, and the runs of each mesh ran in sequence in one allocation.
At $2\times16\times1$, contention bursts of 1 to 3~s per iteration on the shared network affected that allocation and two of the three rounds of an interleaved rerun; Table~\ref{tab:gpt2-a5000} uses the remaining round, in which each run lasted 20 iterations with the first 5 untimed.
On GPT-2 Large, Dion uses rank $r=400$, at which its per-GPU optimizer state equals that of \dobb; on GPT-2 XL, it uses the rank fraction in $\{1/3,1/4,1/8,1/16,1/32\}$ closest to the rank with equal state.
At $2\times16\times1$, where \dobb's blocks are twice as wide as at $1\times32\times1$ and its second-moment reduction also crosses the two replicas, its refresh calls take 98.5~ms, and its amortized call, 23.3~ms, exceeds AdamW's 18.3~ms.

\begin{table}[ht]
\centering
\small
\setlength{\tabcolsep}{4pt}
\caption{GPT-2 optimizer-call times and pooled iteration times on A5000 GPUs, in milliseconds.
The GPT-2 XL runs use activation checkpointing.
Refresh and Non-refresh give \dobb's mean optimizer-call time on iterations with and without an eigenbasis refresh, over three refresh iterations per run and two at $2\times16\times1$.
Amortized is the mean optimizer-call time, for \dobb averaged over one refresh cycle (Appendix~\ref{app:pooled}); the other methods perform the same update on every iteration, so their split entries are empty.
Optimizer-call means cover 30 timed iterations, and 15 at $2\times16\times1$.
Pooled iteration times are derived from the measured means (Appendix~\ref{app:pooled}).
Dion's label gives its rank $r$ or rank fraction $\rho$.
}
\label{tab:gpt2-a5000}
\begin{tabular}{llrlrrrr}
\toprule
 & & & & \multicolumn{3}{c}{Optimizer call (ms)} & Pooled \\
\cmidrule(lr){5-7}
Model & $\DP\times\FS\times\TP$ & GPUs & Method & Non-refresh & Refresh & Amortized & iteration \\
\midrule
GPT-2 Large & $1\times10\times2$ & 20 & AdamW & -- & -- & 11.3 & 7764.5 \\
 & & & Muon & -- & -- & 623.0 & 8376.2 \\
 & & & Dion $r=400$ & -- & -- & 1224.1 & 8977.3 \\
 & & & \dobb & 6.0 & 13.7 & 6.9 & 7760.1 \\
\addlinespace
GPT-2 XL & $1\times32\times1$ & 32 & AdamW & -- & -- & 14.5 & 8059.0 \\
 & & & Muon & -- & -- & 840.9 & 8885.4 \\
 & & & Dion $\rho=1/8$ & -- & -- & 1036.7 & 9081.2 \\
 & & & \dobb & 7.9 & 17.5 & 9.0 & 8053.5 \\
\addlinespace
 & $2\times16\times1$ & 32 & AdamW & -- & -- & 18.3 & 8389.3 \\
 & & & Muon & -- & -- & 742.2 & 9113.2 \\
 & & & Dion $\rho=1/4$ & -- & -- & 1703.1 & 10,074.1 \\
 & & & \dobb & 13.9 & 98.5 & 23.3 & 8394.3 \\
\addlinespace
 & $2\times8\times2$ & 32 & AdamW & -- & -- & 18.7 & 11,648.9 \\
 & & & Muon & -- & -- & 946.5 & 12,576.6 \\
 & & & Dion $\rho=1/3$ & -- & -- & 2523.2 & 14,153.3 \\
 & & & \dobb & 14.2 & 32.9 & 16.3 & 11,646.4 \\
\bottomrule
\end{tabular}
\end{table}

\subsection{GPT-2 on H100 GPUs}
\label{app:distributed-h100}

The two H100 campaigns used H100 80\,GB GPUs.
GPT-2 XL, with 48 layers of width 1600 and 25 attention heads (1.56B parameters), ran at $\DP\times\FS\times\TP=1\times16\times1$ on two nodes of eight GPUs and at $2\times16\times1$ on four such nodes, with micro-batches of 15 sequences and two and one accumulation micro-steps, respectively.
At $\DP=2$, each FSDP group spans two nodes, and the two replicas also all-reduce their gradient shards.
Capacity preemptions split the runs of each GPT-2 XL mesh over two allocations: AdamW, Muon, and Dion $\rho=1/4$ ran in one, and Dion $\rho=1/16$ and \dobb in the other.
GPT-2 3B (60 layers, width 2048, 32 heads; 3.13B parameters), 6B (72 layers, width 2560, 40 heads; 5.79B), and 8B (84 layers, width 2816, 44 heads; 8.14B) ran at $1\times16\times2$, $1\times8\times4$, and $1\times16\times4$ on 32, 32, and 64 GPUs, with micro-batches of 15 sequences and 2, 4, and 2 accumulation micro-steps.
Dion ran at rank fractions $\rho\in\{1,1/3,1/4,1/8\}$ at these three sizes; the 3B run at $\rho=1/3$ came from a separate retry.
Each H100 run lasted 30 iterations, of which the first 10 were untimed, and no H100 run used activation checkpointing.
Table~\ref{tab:gpt2-h100} reports both H100 campaigns.

\begin{table}[ht]
\centering
\small
\setlength{\tabcolsep}{4pt}
\caption{GPT-2 optimizer-call times and pooled iteration times on H100 GPUs, in milliseconds.
Columns follow Table~\ref{tab:gpt2-a5000}.
Optimizer-call means cover 20 timed iterations, and \dobb's Refresh entries cover two refresh iterations per run.
The measured means determine pooled iteration times through the calculation in Appendix~\ref{app:pooled}.
Dion's label gives its rank fraction $\rho$.
$^\dagger$The abstract and Section~\ref{sec:results-fsdp} cite this pair, which has the configuration of the GPT-2 XL loss comparison in Table~\ref{tab:quality}: \dobb with $B=16$ blocks per weight matrix, here one block per FSDP shard, and Dion at $\rho=1/16$.
}
\label{tab:gpt2-h100}
\begin{tabular}{llrlrrrr}
\toprule
 & & & & \multicolumn{3}{c}{Optimizer call (ms)} & Pooled \\
\cmidrule(lr){5-7}
Model & $\DP\times\FS\times\TP$ & GPUs & Method & Non-refresh & Refresh & Amortized & iteration \\
\midrule
GPT-2 XL & $1\times16\times1$ & 16 & AdamW & -- & -- & 8.5 & 1154.3 \\
 & & & Muon & -- & -- & 83.7 & 1229.5 \\
 & & & Dion $\rho=1/4$ & -- & -- & 204.6 & 1350.4 \\
 & & & Dion $\rho=1/16$ & -- & -- & 130.0 & 1275.8 \\
 & & & \dobb & 5.6 & 22.6 & 7.5 & 1153.3 \\
\addlinespace
 & $2\times16\times1$ & 32 & AdamW & -- & -- & 9.3 & 1069.2 \\
 & & & Muon & -- & -- & 82.7 & 1142.6 \\
 & & & Dion $\rho=1/4$ & -- & -- & 232.7 & 1292.6 \\
 & & & Dion $\rho=1/16{}^{\dagger}$ & -- & -- & 124.1 & 1184.0 \\
 & & & \dobb{}$^{\dagger}$ & 5.8 & 36.8 & 9.2 & 1069.1 \\
\addlinespace
GPT-2 3B & $1\times16\times2$ & 32 & AdamW & -- & -- & 49.4 & 2387.7 \\
 & & & Muon & -- & -- & 454.3 & 2792.6 \\
 & & & Dion $\rho=1$ & -- & -- & 3508.4 & 5846.7 \\
 & & & Dion $\rho=1/3$ & -- & -- & 1841.7 & 4180.0 \\
 & & & Dion $\rho=1/4$ & -- & -- & 1385.6 & 3723.9 \\
 & & & Dion $\rho=1/8$ & -- & -- & 1161.9 & 3500.2 \\
 & & & \dobb & 6.7 & 15.0 & 7.6 & 2345.9 \\
\addlinespace
GPT-2 6B & $1\times8\times4$ & 32 & AdamW & -- & -- & 13.9 & 4822.8 \\
 & & & Muon & -- & -- & 2106.8 & 6915.7 \\
 & & & Dion $\rho=1$ & -- & -- & 3407.8 & 8216.7 \\
 & & & Dion $\rho=1/3$ & -- & -- & 1505.7 & 6314.6 \\
 & & & Dion $\rho=1/4$ & -- & -- & 1112.0 & 5920.9 \\
 & & & Dion $\rho=1/8$ & -- & -- & 829.7 & 5638.6 \\
 & & & \dobb & 9.7 & 20.6 & 10.9 & 4819.8 \\
\addlinespace
GPT-2 8B & $1\times16\times4$ & 64 & AdamW & -- & -- & 15.2 & 5337.0 \\
 & & & Muon & -- & -- & 2782.1 & 8103.9 \\
 & & & Dion $\rho=1$ & -- & -- & 4819.4 & 10,141.2 \\
 & & & Dion $\rho=1/3$ & -- & -- & 2081.5 & 7403.3 \\
 & & & Dion $\rho=1/4$ & -- & -- & 1653.1 & 6974.9 \\
 & & & Dion $\rho=1/8$ & -- & -- & 1179.6 & 6501.4 \\
 & & & \dobb & 9.1 & 16.9 & 10.0 & 5331.7 \\
\bottomrule
\end{tabular}
\end{table}

\subsection{Qwen3 FSDP and tensor-parallel series}
\label{app:distributed-qwen}

\paragraph{Setup.}
The Qwen3 series used the matrix shapes of the dense Qwen3-8B and Qwen3-32B models with random initialization \citep{yang2025qwen3}.
Qwen3-8B has 36 layers of width 4096 and MLP width 12288; Qwen3-32B has 64 layers of width 5120 and MLP width 25600.
Both models have eight key-value heads of dimension 128, so the key and value projections have 1024 outputs.
Training read OpenWebText under the GPT-2 tokenizer, whose token indices lie within the Qwen3 vocabulary.
The tokenizer changes token values but not tensor shapes, which determine the computation and communication of each iteration.
Each iteration processed one 1024-token sequence per FSDP rank, a global batch of $\FS$ sequences; the TP ranks of a group shared their sequence.
Each run lasted 80 iterations.
Means use the last 63 of 69 timed iterations, seven complete $T=9$ cycles.
The timed window excludes first-iteration startup: Dion's optimizer compilation and the exact cold-start eigendecompositions of \dobb and \dobbGLtwo.
\dobb and \dobbGLtwo used fp32 second-moment accumulation and one warm orthogonal iteration per refresh.
All runs used H100 GPUs, eight per host.
The runs at $\FS=8$ occupied one host, and the runs at larger $\FS$ spanned hosts.
The Qwen3 campaign compared \dobb and \dobbGLtwo with Dion only and included no Muon or AdamW runs.
Dion ran before \dobb and \dobbGLtwo, with separate allocations.

\paragraph{Block layout.}
With $B=\FS$, the blocks on the 1024-wide key and value axes have width $1024/\FS$.
Longer basis axes receive proportionally wider blocks.
In the Qwen3-8B series, $B=\FS$ makes each output-axis block coincide with one rank's FSDP shard, and \dobbGLtwo requires each rank to hold whole blocks.
Under TP, the Dion-based implementation applies FSDP to the input axis of column-parallel weights and to the output axis of row-parallel weights, so the TP and FSDP shards of each weight lie on different axes.
With $B=\FS$, each block then lies within one shard.
In the Qwen3-32B TP runs, $B$ is 32 or 64 on every matrix, so $B\ge16\cdot\TP$.
\dobb issues no optimizer collective on the TP group.
\dobbGLtwo completes its column norms across TP for row-parallel weights and across FSDP for column-parallel weights; both reductions are the $\mathcal{O}(m)$ entry of Table~\ref{tab:comm-ablations}.
The TP series compares \dobb and \dobbGLtwo with Dion; the Dion-based implementation does not support Muon under TP, whereas our implementation does.

\begin{table}[ht]
\centering
\small
\caption{Qwen3-8B FSDP optimizer-call times and pooled iteration times in milliseconds, with $\TP=1$ and $B=\FS$ blocks per weight matrix.
Non-refresh and Refresh give the mean optimizer-call time on iterations without and with an eigenbasis refresh; Amortized is the mean over the timed window.
Dion performs the same update on every iteration, so its split entries are empty.
Pooled iteration time follows Appendix~\ref{app:pooled}, with shared components averaged over the three runs at each $\FS$.
}
\label{tab:qwen-fsdp}
\begin{tabular}{rlrrrr}
\toprule
 & & \multicolumn{3}{c}{Optimizer call (ms)} & \\
\cmidrule(lr){3-5}
$\FS$ & Method & Non-refresh & Refresh & Amortized & Pooled iteration \\
\midrule
8 & Dion & -- & -- & 159.5 & 422.6 \\
 & \dobb & 75.7 & 170.0 & 86.2 & 349.3 \\
 & \dobbGLtwo & 88.7 & 219.5 & 103.2 & 366.3 \\
16 & Dion & -- & -- & 146.1 & 5317.7 \\
 & \dobb & 26.1 & 331.7 & 60.1 & 5231.7 \\
 & \dobbGLtwo & 28.5 & 336.3 & 62.7 & 5234.3 \\
32 & Dion & -- & -- & 114.8 & 5552.1 \\
 & \dobb & 10.7 & 165.0 & 27.9 & 5465.2 \\
 & \dobbGLtwo & 11.0 & 179.2 & 29.7 & 5467.0 \\
64 & Dion & -- & -- & 138.6 & 5986.8 \\
 & \dobb & 5.5 & 92.1 & 15.1 & 5863.3 \\
 & \dobbGLtwo & 5.5 & 134.1 & 19.8 & 5868.0 \\
\bottomrule
\end{tabular}
\end{table}

\paragraph{FSDP on Qwen3-8B.}
Table~\ref{tab:qwen-fsdp} gives the three-way comparison for Qwen3-8B.
At $\FS=8$, \dobb and \dobbGLtwo have pooled iteration times of 349.3 and 366.3~ms, against 422.6~ms for Dion.
At this degree, where each run occupies one host, Dion's pooled iteration time is 1.21 times that of \dobb.
At $\FS\ge16$, pooled iteration times are 5.2 to 6.0~s, and those of \dobb and \dobbGLtwo are 83.4 to 123.5~ms below Dion's, the differences in their optimizer-call means.
At these degrees, Dion's optimizer call is at most 2.7\% of its pooled iteration time.
Across the Qwen3-8B series, refresh calls take 92.1 to 331.7~ms for \dobb and 134.1 to 336.3~ms for \dobbGLtwo, and both peak at $\FS=16$.

\begin{table}[ht]
\centering
\small
\setlength{\tabcolsep}{4pt}
\caption{Qwen3-32B FSDP and TP optimizer-call times and pooled iteration times in milliseconds, with $B=\FS$ blocks per weight matrix.
Columns and averaging follow Table~\ref{tab:qwen-fsdp}.
}
\label{tab:qwen-tp}
\begin{tabular}{rrlrrrr}
\toprule
 & & & \multicolumn{3}{c}{Optimizer call (ms)} & \\
\cmidrule(lr){4-6}
$\FS$ & $\TP$ & Method & Non-refresh & Refresh & Amortized & Pooled iteration \\
\midrule
32 & 2 & Dion & -- & -- & 1307.7 & 17,664.1 \\
 & & \dobb & 32.8 & 1249.2 & 168.0 & 16,524.4 \\
 & & \dobbGLtwo & 53.0 & 1917.9 & 260.2 & 16,616.6 \\
64 & 2 & Dion & -- & -- & 1287.5 & 18,423.8 \\
 & & \dobb & 14.5 & 634.9 & 83.4 & 17,219.7 \\
 & & \dobbGLtwo & 27.9 & 1007.3 & 136.7 & 17,273.0 \\
64 & 4 & Dion & -- & -- & 797.3 & 14,443.9 \\
 & & \dobb & 8.5 & 560.6 & 69.9 & 13,716.5 \\
 & & \dobbGLtwo & 25.7 & 876.8 & 120.2 & 13,766.8 \\
\bottomrule
\end{tabular}
\end{table}

\paragraph{Tensor parallelism on Qwen3-32B.}
Table~\ref{tab:qwen-tp} reports optimizer-call times and pooled iteration times at $(\FS,\TP)\in\{(32,2),(64,2),(64,4)\}$.
Across \dobb and \dobbGLtwo, optimizer-call means range from 69.9 to 260.2~ms, against Dion's 797.3 to 1307.7~ms, while pooled iteration times are 13.7 to 18.4~s; Dion's optimizer call is at most 7.4\% of its pooled iteration time.
At $\TP=2$, the \dobb and \dobbGLtwo pooled iteration times are 1047.5 to 1204.1~ms below Dion's.
At $\TP=4$, \dobb and \dobbGLtwo have pooled iteration times of 13,716.5 and 13,766.8~ms, against 14,443.9~ms for Dion.
Across the three meshes, Dion's pooled iteration time is 1.05 to 1.07 times that of \dobb.
The Dion-based TP implementation of \dobb reduces its second moments inside the optimizer call rather than from the capture hook, so its optimizer-call times include this reduction.
Refresh calls take 560.6 to 1249.2~ms for \dobb and 876.8 to 1917.9~ms for \dobbGLtwo.

\end{document}